# Efficient Solution to the 3D Problem of Automatic Wall Paintings Reassembly

Constantin Papaodysseus[*, 1], Dimitris Arabadjis[1], Michalis Exarhos[1], Panayiotis Rousopoulos[1], Solomon Zannos[1], Michail Panagopoulos[3], and Lena Papazoglou-Manioudaki[2]

[1] National Technical University of Athens, School of Electrical and Computer Engineering, Iroon Polytechniou 9, 15773, Athens, Greece.

[2] National Archaeological Museum of Greece, Tositsa 1, 10682, Athens, Greece.

[3] Ionian University, Department of Audio and Visual Arts, Corfu, Greece.

[*] corresponding author : Iroon Polytechniou 9, 15773, Athens, Greece, telephone: +30 210 7722329, e-mail: cpapaod@cs.ntua.gr.

**Abstract -** This paper, introduces a new approach for the automated reconstruction-reassembly of fragmented objects having one surface near to plane, on the basis of the 3D representation of their constituent fragments. The whole process starts by 3D scanning of the available fragments. The obtained representations are properly processed so that they can be tested for possible matches. Next, four novel criteria are introduced, that lead to the determination of pairs of matching fragments. These criteria have been chosen so as the whole process imitates the instinctive reassembling method dedicated scholars apply. The first criterion exploits the volume of the gap between two properly placed fragments. The second one considers the fragments' overlapping in each possible matching position. *Criteria 3*, *4* employ principles from calculus of variations to obtain bounds for the area and the mean curvature of the contact surfaces and the length of contact curves, which must hold if the two fragments match. The method has been applied, with great success, both in the reconstruction of objects artificially broken by the authors and, most important, in the virtual reassembling of parts of wall-paintings belonging to the Mycenaic civilization (c.1300 BC.), excavated highly fragmented in Tyrins, Greece.

# 1 Introduction

Many very important archaeological objects are unearthed fragmented, frequently in many hundreds or even thousands of pieces. The problem of reconstructing the initial object from its constituent parts is, as a rule, a very painstaking, tedious and time consuming process. For example, only in Greece, there are thousands of fragmented ancient objects waiting to be reconstructed. A very important class of these objects is the wall-paintings such as those excavated in Mycenae, Tyrins, Akrotiri, Thera, Crete, Pylos, etc. There are numerous wall-paintings of great archaeological value excavated in thousands of fragments that remain broken and non-preserved for tenths of years, exactly because their reconstruction faces serious difficulties.

**1.1 Related works**

There have been various approaches in the treatment of the problem of automated reassembling of fragmented objects. Thus, for example in [4] the reconstruction of 2D fragmented or torn objects is undertaken; the procedure compares the curvature-encoded fragment outlines, at progressively increasing scales of resolution, using an incremental dynamic programming sequence-matching algorithm. In [21] the authors tackle the problem of fragmented pot reconstruction by means of axially symmetric implicit polynomial surface models. The same problem is treated in [14], where earthenware reconstruction is based on average color continuation in contour pixels of adjacent fragments. The pot-shape reconstruction approach of [19] uses functions of the curvature to spot matching of contiguous contour pixels of two fragments. Papers [2], [15], [16], [17], [18], [24] treat the automatic reassembly of torn or shredded documents in a contour-based manner using 2D/3D representations of the documents.

Approaches to the problem of fragmented objects' 3D reconstruction usually incorporate elements from surface matching (e.g. the ICP algorithm introduced in [1] or point-by-point approaches like the "Generalized Hough Transform" [7]) and schemes from pattern recognition in order to determine the proper sequence of optimal surface alignments that possibly solve the reconstruction problem. Namely, in [22] the authors initially reduce the dimensionality of the surface alignment problem by constraining the surface normals' alignment and then search for optimal pairwise matches by a special kind of random sample consensus (RANSAC) scheme. In [8] and [9] automatic 3D reconstruction is dealt via point by point distances (z-buffer) between given mutually visible facets of the object's fragments. The optimal alignment between adjacent fracture facets is spotted via simulated annealing optimization. In [5], the authors introduce a 3D reconstruction method based on fragments' surface features computed via centered multi-scale local integrals. Potential fracture surfaces are spotted via a graph-cuts based segmentation algorithm. Then a solution to fragments' reassembly is determined via feature-based global registration for pairwise matching of fragments, and simultaneous constrained local registration of multiple fragments. The approach introduced in [23] differs from the previous ones in the sense that the proposed method for fragmented objects' 3D reassembly is not a feature-based one, but relies on the action of an iterative process over a dense binary tree structure attached to the fracture facets' points. Namely, the authors employ a transformation between pairs of points on adjacent fracture surfaces in order to define binary relations between these surfaces. In order to determine the maximal set of neighboring points that satisfy the same binary relation the authors employ an hierarchical clustering algorithm which iteratively acts over the clusters binary tree in a "region-growing" manner thus decreasing its density. The problem of fragmented wall-paintings 3D reassembly is discussed on [3]. The authors present an inexpensive system for 3D fragment information acquisition and processing. The acquisition system requires minimal supervision, so that a single, non-expert user can scan at least 10 fragments per hour. The system is applied to Akrotiri, Thera wall-paintings. In [10], the problem of fragmented wall-paintings reconstruction is treated, in the case where the only available information is the set of 2D fragments' images; the approach is effective but it suffers from the intrinsic restriction that there is no available three-dimensional information of the constituent parts.

## 1.2 Contribution of the present work

In the present paper a methodology and a related information system are presented that tackle the problem of automated reconstruction of an arbitrary fragmented object, with the only restriction that one of the fragments' surface is plane or near to plane, as in the case of the wall-paintings [11]. The goal of the present work is to propose a system that offers a practically unique solution, as far as fragments matching is concerned. In other words, the introduced system does not model the contact surfaces, nor it considers the exact shape and positions of the fracture facets; it rather tries to imitate the process the dedicated scholars instinctively follow, in their attempt to reassembly the wall-painting. In addition, the methodology takes into account the unavoidable wear the unearthed archaeological fragments suffer and determines extreme cases concerning the geometry of adjacent fragments by means of calculus of variations. We have applied the introduced methodology and the related system a) in an artificial test case and b) in the actual, very important case of prehistoric wall-paintings reconstruction with great success. In both cases the constituent parts have been correctly matched and the proper matching position between actually adjacent fragments has been uniquely spotted; in the case of the wall – paintings, the previous statement about correctness of the results expresses the fact that dedicated scholars (archaeologists and conservators) fully agreed with the matches proposed by the system.

**1.3 A brief description of the introduced approach**

First, we perform a 3D scanning of the available fragments (Section *5.1*). Next, for each fragment image we automatically spot its upper near – to – plane surface (Sect. 2.1 and Appendix 1). We also determine the axis of least moment of inertia (fragment's "central axis"), normal to the upper plain surface (Sect. 2.2) and we rotate all fragments so that their central axes are parallel to the z-axis. Next, we generate a large set of rotated versions of each fragment, by rotating it around its central axis by a small angular step $\delta\theta$ (Sect. 2.3). This action takes place only once for each fragment.

In order to test if two fragments A, B match, we place each rotated version of B properly adjacent to A (Sect. 3.1) and we define possible contact surfaces between them. At each position tested for matching we apply four criteria, 3 necessary ones and 1 sufficient. Specifically,

A) We check if the relative lengths of the contact curves in the common upper surfaces of the fragments are acceptable according to a new proposition stated in Sect. 3.5.

B) We examine if the area of the "contact surfaces" of the two fragments at the specific relative position satisfies the Theorem stated and proved in Sect. 3.4, whose exact content is derived by means of calculus of variations.

C) We look for possible overlapping between the two fragments at the position in hand acting as described in Sect. 3.3.

D) If all, sequentially applied criteria, (A), (B), (C), are satisfied, then for the relative placement of fragments A and B, we define a proper 3D domain, between the two fragments and we compute its volume. If this volume is smaller than a properly predefined threshold, then the relative placement of the two fragments is characterized as a matching one.

We would like to point out that, ostensibly, testing for matching all rotated versions of each fragment is a rather cumbersome procedure. In fact, we could have evaluated the rejection criteria of Sect. 3.4 and 3.5 so as to be rotational invariant. For example one could have used an ensemble of chains of equal length, or sectors of constant dihedral angle in both fixed and rotating fragments to evaluate this criterion, without applying rotation to any fragment. However, this could reduce the precision of the system in determining the optimal matching surfaces, while our approach is strongly oriented towards achieving precision, rather than speed. Evaluating all rotated versions of each fragment is essential and necessary in order to compute the volume enclosed by the adjacent fracture surfaces, while at the same time we avoid overlapping. Nevertheless, as already mentioned, the rejection criteria developed by means of calculus of variations drastically reduce the complexity of the calculations and correspondingly speed up the whole process. The flowchart of the whole matching procedure is presented in Fig. 1.

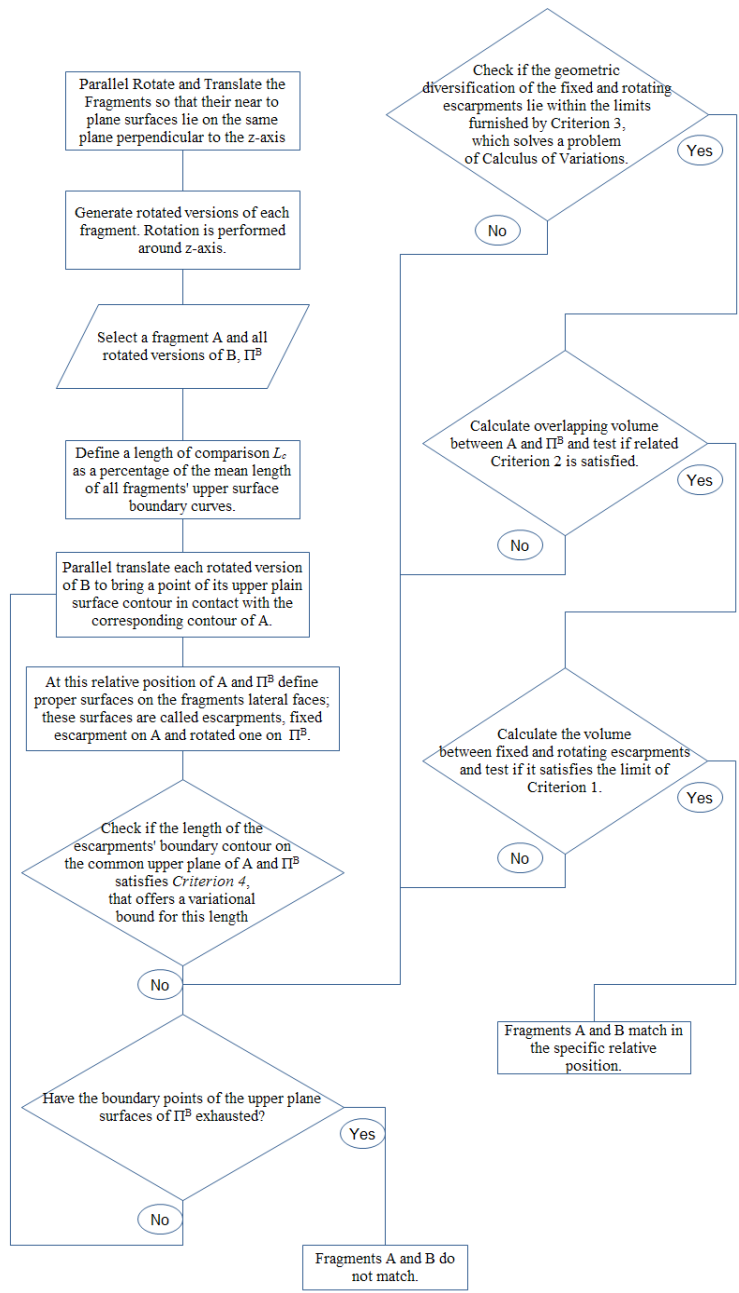

**Fig. 1.** Block diagram of the whole procedure proposed for testing if there is a proper matching position between two fragments.

## 2 Preliminary Fragments Processing

### 2.1 Defining the upper and bottom planes of the fragment and its lateral surface

First, we spot the upper, frequently painted, plain surface of the wall painting fragment, by means of the method described in Appendix 1. Let $A^{US}x+B^{US}y+C^{US}z+D^{US} = 0$ be the equation of the upper plane. We parallel translate it along its normal vector to the direction in which the plane has nonempty intersection with the fragment boundary surface. This parallel translation continues until this intersection becomes empty. At this point we move backwards until the intersection of the plain with the fragment surface forms a closed curve defining an area equal or just grater than a threshold area $\alpha^{\min}$. The corresponding plane $A^{US}x+B^{US}y+C^{US}z+D^{BS} = 0$ is considered to be the bottom plane of the fragment in hand. We emphasize that this plain is an auxiliary one and changes according to the considered fragment depth; there is no demand that the back side of the wall-painting fragments belong to the same plane.

Finally, we define the lateral surface of the fragment to be the maximal connected subset of the fragment surface lying between fragment's upper and bottom planes (see Figure 2).

### 2.2 Defining the central axis of each fragment

In the following we will consider the fragment to be the shape bounded by the intersection of the upper plane with the fragment surface, the intersection of the bottom plane with the fragment surface and the lateral surface of the fragment. We will also treat this shape as a homogenous 3D body.

We determine the axis that passes through "the center of gravity" of this body which is parallel to the vector $\vec{n}^{US} = \left(A^{US}, B^{US}, C^{US}\right)$ and hence vertical to the upper and bottom planes. We call this, the central axis of the fragment. In addition, we rotate all fragments by proper Euler angles so as the central axis coincides with $z$-axis.

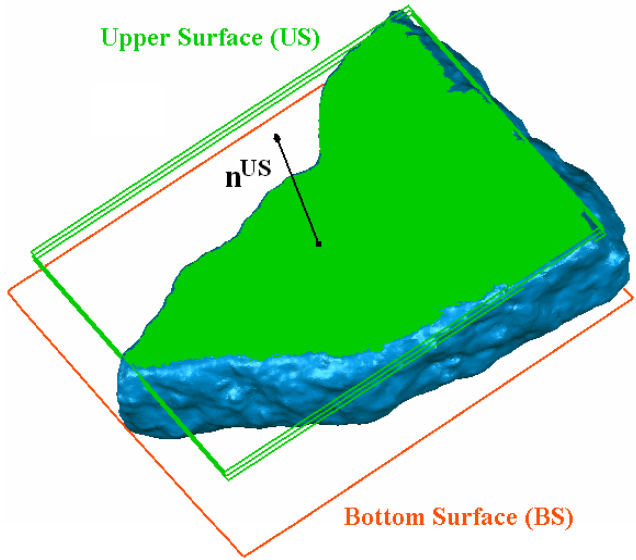

**Fig. 2.** Preliminary fragments processing. Fragment's upper plane surface and the corresponding auxiliary shape are depicted with the green surface points and the green frame correspondingly. Fragment's central axis is along **n**$^{US}$ and passes from the centroid of the fragment. Fragment's bottom plane is depicted with the red frame, it is parallel to the upper plane and it is determined as described in Sect. 3.1.

### 2.3 Generating rotated versions to cope with fragments' random orientation

In our attempt to find if two fragments actually match, we must cope with the arbitrariness in orientation of them. To circumvent this difficulty we will generate rotated versions of all available fragments. Namely, we rotate each fragment around its central axis (z-axis), by a very small angular step $\delta\theta=1^o$, thus obtaining the sequence of angles $\theta_i = \delta\theta \cdot i$, $\theta_i \in [0, 2\pi)$ and all corresponding fragment's rotating versions. The ensemble of all rotated versions of an arbitrary fragment, say the *R*-th, form a set $\Pi^R$.

# 3 Analysis of the Method for 3D Wall Paintings Reconstruction – Four New Matching Criteria

### 3.1 Employing the notions of "fixed" and "rotating" fragments pair. Proper relative placement of the fragments and definition of the sub-domain of contact.

Suppose that two fragments are given and that one wants to decide if they match and if yes, where they match, using the entire three – dimensional information. In order to achieve this, one first proceeds as follows:

A first, arbitrarily chosen fragment, called "fixed", is placed in a system of reference, so that its central axis lies on the *z*-axis. Next, one considers a length of comparison measured in pixels, say $L_C$; we use the term pixels to denote the points of the digital contour of fragment's upper plane surface. At first, one considers a group of $L_C$ consecutive pixels starting from pixel #1 of the contour curve $C^F$ of the fragment plane surface. These contour pixels are called "fixed chain", which is denoted as $\Gamma^F_1$, where F stands for "fixed" and 1 for the starting pixel of the contour $C^F$.

Subsequently, we place the second fragment to the same Cartesian system, once more so that its central axis is parallel to the $z$-axis. This second fragment has an arbitrary orientation around its central axis and an arbitrary position as well. We have previously described how rotated versions $\Pi^R$ of the second fragment are generated.

At this point, for each angle $\theta_i$, we parallel translate the rotated fragment to the vicinity of the fixed fragment as follows:

Suppose that the contour curve $C^R$ of the plane surface of the rotated fragment consists of M pixels. Then we parallel translate it so as the first pixel of $C^F$ and the last pixel of $C^R$ coincide. Next, we define the "terminal barrier surface" $B^T_1$ as the plane passing from the last pixel of the fixed chain, which is parallel to the central axis of the fixed fragment and perpendicular to the straight line that joins the first and the last pixel of the fixed chain (Figure 3(b)).

We create a rotating chain $\Gamma^R_{1,M}$ moving on $C^R$ counter clockwise staring from pixel #M and ending on the intersection of the terminal barrier surface with $C^R$, if any. As we move counter clockwise on $C^R$ increasing the length of $\Gamma^R_{1,M}$, if its number of pixels exceeds a proper threshold $L^{EX}$, then we keep a flag that we reject the matching position in hand. A variational estimation of $L^{EX}$ will be given in Sect. 3.5.

Subsequently, we define the notions of "fixed escarpment" and "rotated escarpment". In the process of doing so, we define first the "starting barrier surface" $B^S_1$, as the plane passing from the first point of the fixed chain, which is parallel to the terminal barrier surface $B^T_1$. We also define a common bottom plane for the pair A, B which is the bottom plane of either A, or B, closer to their common upper surface. Then, we define the "fixed escarpment", $E^F_1$, as the surface lying on the lateral boundary surface of the fixed fragment enclosed by the upper plane surface of the fragment, the common bottom plane surface, as well as the initial and terminal barrier planes. Similarly we define the notion of "rotated escarpment", $E^R_{1,M}$, as the lateral boundary surface of the rotated fragment, confined by its upper plane surface, the bottom plane and the initial and terminal barrier planes see (Fig. 3(c)).

We repeat the previous process by changing the starting point of the rotating chain moving from pixel #M to pixel #1 of $C^R$, thus forming an ensemble of rotated chains and rotated escarpments $\Gamma^R_{1,m}$, $E^R_{1,m}$, $m$ = M,M-1,… Finally, we generate two ensembles of fixed chains and fixed escarpments by moving the starting point '$k$' of the fixed chain along $C^F$ namely $\Gamma^F_k$, $E^F_k$, $k$=1,2,…

We stress that for two given fragments, the precise form of the fixed and rotating escarpments $E^F_k$, $E^R_{k,m}$ depends on: the first and last point of $\Gamma^F_k$, $\Gamma^R_{k,m}$, the position of their bottom planes in respect with their common upper one and the angle of rotation $\theta_i$. We would like also to point out that if, for a certain position of the fixed escarpment, there is no intersection between the terminal barrier surface $B^T_k$ and the rotating fragment, then no rotating escarpment is defined. For the system, this means that there is no matching in the specific relative position of fragments A, B.

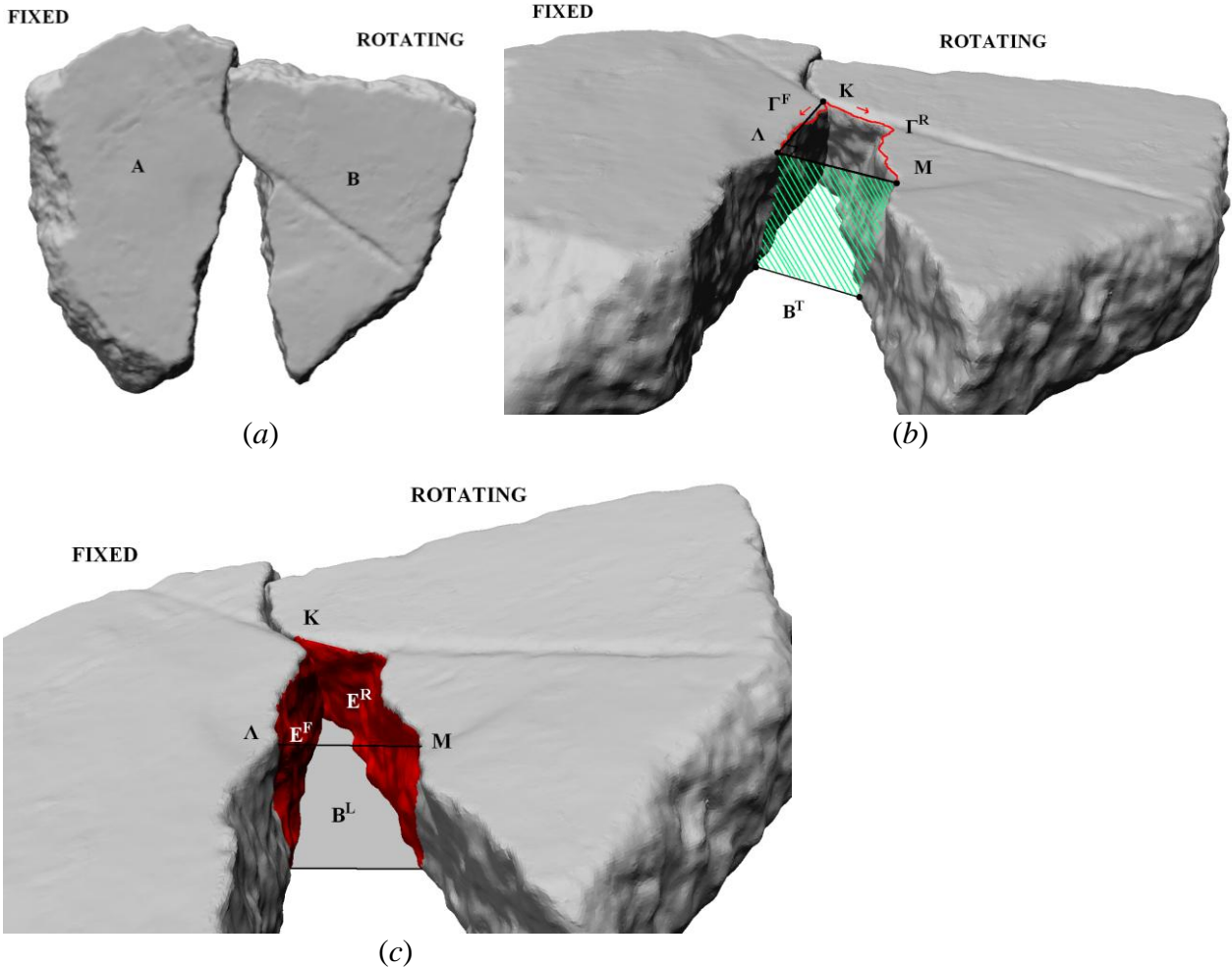

**Fig. 3**: Fixed and Rotating fragments' pair relative placement and definition of the contact sub-domain.
(*a*): Proper relative placement of a pair of fragments tested for matching.
(*b*): Fragments pair's terminal barrier plane and the obtained contact curves, $\Gamma^F$ for the fixed chain and $\Gamma^R$ for the rotating one.
(*c*): Determination of the fragments pair contact surfaces, $E^F$ on the fixed and $E^R$ on the rotating fragment.

## 3.2 A first matching criterion: The volume of a properly chosen 3D domain between adjacent fragments

We consider any two fragments A, B and all related existing pairs a) of fixed and rotating chains ($\Gamma^F_k$, $\Gamma^R_{k,m}$), b) of fixed and rotating escarpments ($E^F_k$, $E^R_{k,m}$), c) of starting and terminal barrier surfaces ($B^S_k$, $B^T_k$). In addition, we place the two fragments in the same frame of reference as described in Sect. 3.1, so as their upper plane surfaces lie on the same plane vertical to the *z*-axis.

Next, let $V_{k,m}$ be the closed domain bounded by a) the fixed and rotating escarpments of the same pair ($E^F_k$, $E^R_{k,m}$), b) starting and terminal barrier surfaces of the same pair ($B^S_k$, $B^T_k$), c) the common upper plane of the two fragments and d) the bottom plane of A or B which is nearest to the common upper plane (see Fig. 4). We compute the volume $\tau_{k,m}$ of all these closed domains $V_{k,m}$; if $\tau_{k,m}$ is smaller than a predefined threshold $\tau^T$, then we consider the specific position as a possible matching position of the two fragments A, B.

We note that the proper choice of $\tau^T$ depends on the comparison length $L_C$, the distance of the common bottom and upper planes and the gap between two actually matching fragments we are willing to accept.

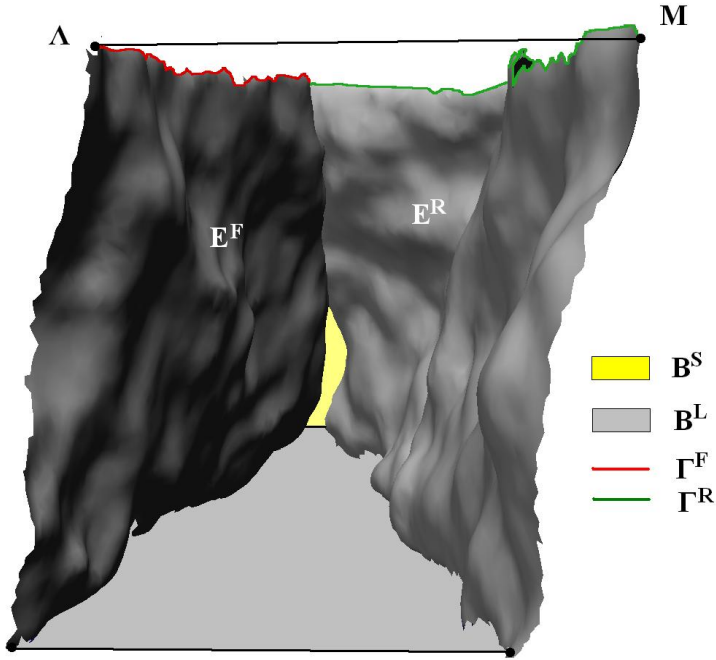

**Fig. 4**: Isolation of the subdomain $V_{k,m}$ whose volume is checked via *Criterion 1*

## 3.3 Second matching criterion: Prohibiting considerable overlapping between adjacent fragments

*Matching Criterion 2: Prohibiting local and overall overlapping in the contact domain $V_{k,m}$*

In the ideal case, where two fragments A, B actually match, there should be no overlapping at all between the corresponding fixed and rotating escarpments. However, in practice, due to unavoidable inaccuracies of digital representation of the fragments, one may expect slide overlapping between these escarpments even in the case of perfect physical matching. Hence, if at a certain position of fragments A, B *Criterion 1* is satisfied, then we demand that the local overlapping between the fixed escarpment $E^F_k$ and the rotating one $E^R_{k,m}$ is smaller than an acceptable threshold $\tau^O$. If overlapping is greater than $\tau^O$, still we must take into account the arbitrariness in the parallel translation of fragment B. Consequently, we proceed as follows in order to remove this arbitrariness: For any point '*p*' of the rotating escarpment, we consider the intersection of the plane vertical to *z*-axis passing from '*p*' with the fixed escarpment; we find the minimum distance $\mu^p$ of '*p*' from all points of this intersection and let $\vec{\mu}^p$ be the corresponding vector. Among all those $\vec{\mu}^p$ we spot the one of maximum length, say $\vec{\mu}$. We parallel translate rotating fragment B by $-\vec{\mu}$ from the position it acquired in the process of Sect. 3.1 and we recalculate overlapping volume. If the overlapping volume is smaller than $\tau^O$ and *Criterion 1* is still satisfied, then we take into account the overall overlapping of the two considered fragments. Namely, we allow for an overall overlapping between the digital representations of the specific fragments and we demand the overlapping volume to be smaller than a proper threshold $\tau^{FO}$. Evidently, the exact value of this threshold depends on the volume of the involved fragments and the quality of the employed method for digital representation. After extensive tests we found out that a very good value for $\tau^{FO}$ is a very small percentage 0.25% of the minimum volume of fragments A, B, for the employed digital representation method.

**3.4 Employing principles of Calculus of Variations to define a third matching criterion: Associating the geometry of the contact surfaces with the maximum allowed volume of the domain $V_{k,m}$**

We will state now a third criterion, which is rather a necessary condition than a sufficient one. However, this criterion, acting as a "matching rejection filter", has drastically accelerated the performance of the introduced 3D automatic reconstruction system. The basic underlying concept may be described as follows: Suppose that at a certain position of fragments A, B perfect matching occurs, in the sense that $\tau_{k,m}$ is zero or equivalently escarpments $E^F_k$, $E^R_{k,m}$ coincide. In this ideal case, at every point of the escarpments the unit vectors normal to them $\vec{n}_F$ and $\vec{n}_R$ are opposite. Since, in practice, we must take into account presence of unavoidable gaps between matching fragments, due the process that fragmented the initial object and/or due to wear, we will definitely meet with deviations from this ideal situation. All the same a corresponding necessary condition must always hold: In fact, one may intuitively expect that if there is a considerable overall diversification of the unit normal vectors $\hat{n}$ throughout the boundary surface of the domain $V_{k,m}$, then its volume $\tau_{k,m}$ can not be satisfactorily small; in other words, in this case, *Criterion 1* will not be satisfied. A rigorous formulation of this intuitive statement is described below.

*Matching Criterion 3* – Consider that fragments A, B are placed as described in Sect. 3.1 and let the volume $\tau_{k,m}$ be less than equal to the grater acceptable value $\tau^T$. Then the integral of angles defined via the formula $\mu = \oint_{\partial V_{k,m}} \arctan\left(\frac{\vec{n} \cdot \hat{j}}{\vec{n} \cdot \hat{i}}\right) dS$ satisfies inequality (4.1), with $r_T$, $r_0$ to be the distances of $B^S_k$, $B^T_k$ from the central axis of $E^F_k$, $\Delta\theta = \theta_T - \theta_0$, to be the dihedral angle that encompasses $E^R_{k,m}$, $\Delta z = T - S$ the distance between the fragments' common upper and lower planes and $r(\theta,T)$, $r(\theta,S)$ the upper and the lower boundary curves of $E^R_{k,m}$.

Equivalently if, for the current position of fragments A, B, the mean angle of the normal vectors of $\partial V_{k,m}$ exceeds the upper bound (4.1), then the position in hand is not an actual matching one. This rejecting criterion has been obtained by solving the variational problem stated below.

*3.4.1 A rigorous statement of the problem using Calculus of Variations*

Let a domain $U$ with boundary surface $\partial U$ represent the gap between two adjacent fragments and their lateral surface parts in this position, respectively. The integral of angles of the unit vectors normal to $\partial U$ is expressed via quantity

$$\mu = \oint_{\partial U} \arctan\left(\frac{\vec{n} \cdot \hat{j}}{\vec{n} \cdot \hat{i}}\right) dS .$$

Then the problem in hand is expressed as follows: find the extremes of quantity $\mu$ given that the volume $V$ of domain $U$ is bounded by $0 \leq V \leq V_M$. Hence, we can formulate the problem by using the Lagrangian integral

$$\iint_{E^R} f\left(\vec{x}, \frac{\partial \vec{x}}{\partial u}, \frac{\partial \vec{x}}{\partial v}, u, v\right) du dv = \mu - \lambda \cdot V$$

where $(u,v)$ are the independent variables of the surface $\partial U$, $\vec{x}(u,v)$ is the position vector of an arbitrary point of $\partial U$, $\frac{\partial \vec{x}}{\partial u}, \frac{\partial \vec{x}}{\partial v}$ the corresponding partial derivatives of $\vec{x}$. Hence the problem is transformed into finding surface $E^R$ (the rotating escarpment) terminated on barrier plane surfaces $B^S$, $B^T$ and common upper and bottom planes $C$, $D$, such that $\delta \iint_{E^R} f\left(\vec{x}, \frac{\partial \vec{x}}{\partial u}, \frac{\partial \vec{x}}{\partial v}, u, v\right) du dv = 0$.

The configuration suggests use of cylindrical coordinates and expression of $\partial U$ with $\theta$ and $z$ being the independent variables. In fact, let $\vec{x}(r, \theta, z)$ be the position vector of an arbitrary point lying on $\partial U$ in cylindrical coordinates; the beginning of vector $\vec{x}$, i.e. the reference point, is considered to be in the internal of domain $U$ to ensure uniqueness in the value of $\vec{x}$. Clearly $\vec{x}(\theta, z) = (r(\theta, z)\cos(\theta), r(\theta, z)\sin(\theta), z)$.

In Cartesian coordinates, the volume $V$ of domain $U$ is $V = \int_U dxdydz$. By applying Stokes Theorem we obtain $V = \frac{1}{3} \oint_{\partial U} \vec{x}(\theta, z) \cdot \vec{n}(\theta, z) d\theta dz$. Since $r$ and $\vec{x}$ are functions of $\theta$ and $z$ the integral that expresses the volume enclosed by $E^R$, $B^S$, $B^T$, and the $xy$-plane is written

$$V = \frac{1}{3} \int_S^T \int_{\theta_0}^{\theta_T} r(\theta, z)^2 - zr(\theta, z) \cdot \frac{\partial}{\partial z} r(\theta, z) d\theta dz \qquad (4.0).$$

The problem now may be stated in a strict manner as follows: "Extremize $\mu$ under the constrains that $V \leq V_M$, where $V_M$ constant, and the barrier surfaces $B^S \subset \partial U$, $B^T \subset \partial U$ are known and fixed planes; the fixed escarpment $E^F$, as well as the starting curve of the rotating escarpment $E^R$ are also known". In other words, the problem is to determine the rotating escarpment that maximizes $|\mu|$ when $V \leq V_M$. A general solution to this problem is given in the following:

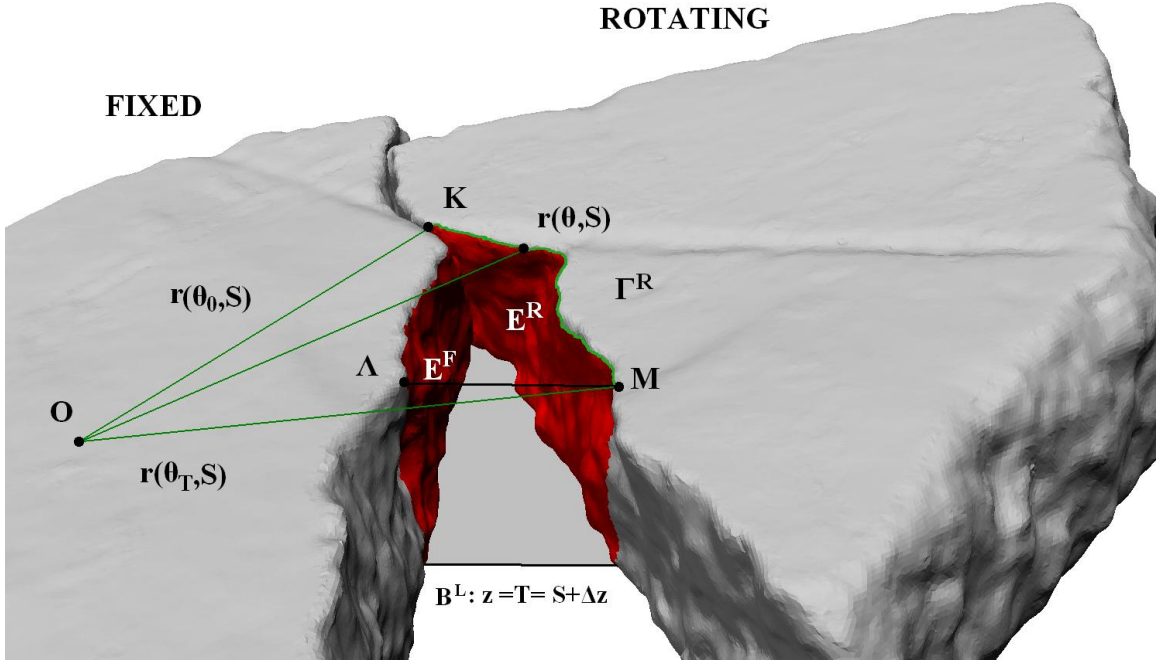

**Fig. 5:** Schematic interpretation of *Criterion 3* notation. Function $r(\theta,S)$ corresponds to the distance of the points of $\Gamma^R$ from the central axis of the fixed fragment, while $r(\theta,T)$ corresponds to the distance of the points belonging to the intersection of $E^R$ with $B^L$ from this axis. *Criterion 3* employs these functions in order to determine the maximum angular discrepancy between the normals of $E^F$ and $E^R$ as equation (4.1) implies.

**Theorem** Consider a fixed rectangular tube $T$, i.e. a rectangular parallelepiped of infinite length, that delimits a 2D domain $E^F$ on a certain surface $S^F$. Let $\Gamma^F$ be the intersection of $E^F$ with $T$; let moreover $\Gamma^R$ be a given piece-wise smooth curve on $T$ that does not cross $\Gamma^F$. Hence, consider the domain $\Omega$ that $\Gamma^F$ and $\Gamma^R$ define on $T$; evidently area($\Omega$)≥0. A fixed axis, say $z$, is also given which is vertical to 2 parallel planes of $T$, non-intersecting $\Omega$ and $E^F$, and placed on the other side of $\Gamma^R$ with respect to $\Gamma^F$. Consider also the cylindrical coordinates system $(r,\theta,z)$ having $z$ as axis. In this system $\Gamma^F$ is enclosed by a dihedral angle starting at $\theta_0$ and ending at $\theta_T$. Consider any piece-wise smooth surface $E^R$ bounded by $\Gamma^R$, with the only restriction that the volume $V$ of the 3D domain enclosed by $T$, $E^F$ and $E^R$ has an upper bound $V \leq V_M$.

Then, the quantity defined via the formula $\mu = \int_{E_R} \arctan\left(\dfrac{\vec{n}\cdot\hat{j}}{\vec{n}\cdot\hat{i}}\right) dS - \int_{E_F} \arctan\left(\dfrac{\vec{n}\cdot\hat{j}}{\vec{n}\cdot\hat{i}}\right) dS$

satisfies the inequality

$$\left|\frac{\mu}{\Delta\theta\Delta z}\right| \leq \left|\ln\left(\frac{r_T}{r_0}\right)\right| + \left|\frac{\Delta\theta}{2} + \frac{\Delta z}{6\tau^T}\cdot\left[-r_0^2 + \frac{1}{\Delta\theta\Delta z}\int_{\theta_0}^{\theta_T}\left(T\cdot r(\theta,T)^2 - S\cdot r(\theta,S)^2\right)d\theta\right]\right| \quad (4.1)$$

where $r_T$, $r_0$ are the lengths of the radii of the points of $\Gamma^R$, which correspond to $\theta_0$ and $\theta_T$ respectively, $\Delta\theta = \theta_T - \theta_0$, $\Delta z = T - S$ is the distance of the two planes of $T$ that are perpendicular to z and $r(\theta,T)$, $r(\theta,S)$ are the radii of the parts of $\Gamma^R$ for $z=T$ and $z=S$ respectively.

*Proof*

The Lagrangian integral that describes the proposed problem reads

$$I = \int_S \int_{\theta_0}^{\theta_T} f(r, r_\theta, r_z) d\theta dz = \int_S \int_{\theta_0}^{\theta_T} \arctan(\varphi) d\theta dz - \frac{\lambda}{3} \int_S \int_{\theta_0}^{\theta_T} r(\theta,z)^2 - zr(\theta,z) \cdot r_z(\theta,z) d\theta dz + \lambda V_M$$

where $\varphi = \dfrac{r\sin(\theta) - r_\theta \cos(\theta)}{r_\theta \sin(\theta) + r\cos(\theta)}$, $r_\theta = \dfrac{\partial r}{\partial \theta}$, $r_z = \dfrac{\partial r}{\partial z}$.

Variation $r(\theta,z) = r^{OPT}(\theta,z) + \eta(\theta,\zeta)$ of optimal function $r^{OPT}$ offers first order variation of $I$

$$\delta I = \int_S \int_{\theta_0}^{\theta_T} \left[ \frac{\partial f}{\partial r} - \frac{\partial}{\partial \theta}\left[\frac{\partial f}{\partial r_\theta}\right] - \frac{\partial}{\partial z}\left[\frac{\partial f}{\partial r_z}\right] \right] \eta d\theta dz + \int_S \left[\frac{\partial f}{\partial r_\theta} \eta\right]_{\theta_0}^{\theta_T} dz + \int_{\theta_0}^{\theta_T} \left[\frac{\partial f}{\partial r_z} \eta\right]_S^T d\theta.$$

In this case the Euler-Lagrange equation is

$$\frac{\partial f}{\partial r} - \frac{\partial}{\partial \theta}\left[\frac{\partial f}{\partial r_\theta}\right] - \frac{\partial}{\partial z}\left[\frac{\partial f}{\partial r_z}\right] = 0 \qquad (4.2);$$

with $f(r, r_\theta, r_z) = \arctan(\varphi) - \dfrac{\lambda}{3} \cdot r \cdot (r - zr_z) + \lambda \dfrac{V_M}{\Delta\theta\Delta z}$, where $\Delta\theta = \theta_T - \theta_0$, $\Delta z = T-S$.

Also, problem's boundary conditions read $\left[\dfrac{\partial f}{\partial r_\theta}\right]_{\theta=\theta_0,\theta_T} = 0$, $f|_{\theta=\theta_0} = 0$. After calculating the partial derivatives of $f$ and multiplying with $r$, (4.2) reads

$$\frac{\partial}{\partial \theta}\left[\frac{r^2}{r^2 + r_\theta^2}\right] = \lambda r^2 \qquad (4.3).$$

By integrating this equation we obtain

$$\frac{r^2}{r^2 + r_\theta^2} + c(z) = \lambda \int_{\theta_0}^{\theta_T} r^2 d\theta \Rightarrow \int_S^T c(z) dz \le \lambda \int_S \int_{\theta_0}^{\theta_T} r^2 d\theta dz \le \int_S^T c(z) dz + \Delta z \qquad (4.4).$$

At this point, we will try to associate the middle integral in (4.4) with the volume $V$ of $U$. In fact, using (4.0), we obtain

$$\frac{1}{2} \int_S \int_{\theta_0}^{\theta_T} r^2 d\theta dz = V + \frac{1}{6} \int_{\theta_0}^{\theta_T} \left(T \cdot r(\theta,T)^2 - S \cdot r(\theta,S)^2\right) d\theta \qquad (4.5).$$

Substituting (4.5) into (4.4) we get

$$\frac{1}{2}\int_S^T c(z)dz \le \lambda V + \frac{\lambda}{6}\int_{\theta_0}^{\theta_T}\left(T\cdot r(\theta,T)^2 - S\cdot r(\theta,S)^2\right)d\theta \le \frac{1}{2}\int_S^T c(z)dz + \frac{\Delta z}{2} \quad (4.6).$$

Letting $0 \le V \le V_M$, (4.6) gives bounds for $\lambda$ via the inequality

$$|\lambda| \le \frac{\Delta z}{V_M} \quad (4.7).$$

Below we will use this upper bound of $|\lambda|$ to compute the extremes for the bounds of the integral of angles.

First, we define $a = \frac{r_\theta}{r}$. Under this suggestion the Lagrangian can be written

$$f = \theta - \arctan(a) - \frac{\lambda}{3}\cdot r \cdot (r - zr_z) \quad (4.8).$$

In addition, using the previous definition of $\alpha$, equation (4.3), after integration with respect to $\theta$, reads

$$\frac{a}{1+a^2} - \arctan(a) = \frac{\lambda}{2}r^2 - g(z) \quad (4.9).$$

There result the inequalities $-|a| - \frac{\lambda}{2}r^2 + g(z) \le \arctan(a) \le |a| - \frac{\lambda}{2}r^2 + g(z)$, which after integration becomes

$$\left|\int_{\theta_0}^{\theta_T}\arctan(a)d\theta - \Delta\theta g(z) + \frac{\lambda}{2}\int_{\theta_0}^{\theta_T}r^2 d\theta\right| \le \int_{\theta_0}^{\theta_T}|a|d\theta \quad (4.10).$$

Additionally the equation (4.3) suggests that $|a| = \frac{|r_\theta|}{r}$ is strictly monotonous with respect to $\theta$, hence $\int_{\theta_0}^{\theta_T}|a|d\theta = \left|\int_{\theta_0}^{\theta_T} a d\theta\right| = \left|\ln\left(\frac{r(\theta_T,z)}{r(\theta_0,z)}\right)\right| = \Lambda(z)$. Therefore inequality (4.10) gives

$$\left|\int_{\theta_0}^{\theta_T}\arctan(a)d\theta - \Delta\theta g(z) + \frac{\lambda}{2}\int_{\theta_0}^{\theta_T}r^2 d\theta\right| \le \Lambda(z) \quad (4.11).$$

Now, we express the sought for integral of angles $\mu$ by means of $\alpha$; in fact, $\mu = \frac{\Delta(\theta^2)}{2}\Delta z - \int_S^T\int_{\theta_0}^{\theta_T}\arctan(a)d\theta dz$, $\Delta(\theta^2)=\theta_T^2-\theta_0^2$, and using (4.11) we obtain

$$\left| \frac{\Delta(\theta^2)}{2}\Delta z - \mu - \Delta\theta \int_S g(z)dz + \frac{\lambda}{2}\int_S \int_{\theta_0}^{\theta_T} r^2 d\theta dz \right| \leq \int_S \Lambda(z)dz \qquad (4.12).$$

The last step is the computation of the integral $\int_S g(z)dz$. This will be accomplished by exploitation of the initial conditions of the Euler-Lagrange equations. Indeed, the initial barrier surface, as well as the initial barrier curve in the rotating fragment may be considered known and fixed, in which case the zero first order variation imply that $f|_{\theta=\theta_0} = 0$. By substituting (4.9) into this equation and after integrating with respect to 'z' we obtain

$$\int_S g(z)dz = \theta_0 \Delta z + \lambda \frac{V_M}{\Delta\theta} + \frac{\lambda}{6}\cdot\left(Tr_0(T)^2 - Sr_0(S)^2\right) - \int_S \frac{r_0^\theta r_0}{r_0^{\theta 2} + r_0^2} dz \qquad (4.13),$$

where $r_0(z) = r(\theta_0, z)$, $a_0 = \frac{r_0^\theta}{r_0}$, $r_0^\theta = r_\theta(\theta_0, z)$. But the demand that $E^R$ starts at $\theta_0$ on barrier plane $B^S$ and ends at $\theta_T$ on barrier plane $B^T$ implies that

$$\left.\frac{\partial f}{\partial r_\theta}\right|_{\theta=\theta_B} = 0 \Rightarrow \frac{r(\theta_B, z)}{r_\theta(\theta_B, z)^2 + r(\theta_B, z)^2} = 0, \text{ where } \theta_B = \theta_0 \text{ or } \theta_T.$$

Hence

$$\frac{1}{r_\theta(\theta_0, z)} \to 0, \quad r(\theta_0, z) = r_0 \qquad (4.14),$$

$$\frac{1}{r_\theta(\theta_T, z)} \to 0, \quad r(\theta_T, z) = r_T \qquad (4.15).$$

Using relations (4.14) and (4.15) we obtain the simplified form for (4.13)

$$\frac{r_0^\theta r_0}{r_0^{\theta 2} + r_0^2} \xrightarrow{\frac{1}{r_0^\theta} \to 0} 0 \Rightarrow \int_S^T g(z)dz = \theta_0 \Delta z + \lambda \frac{V_M}{\Delta\theta} + \frac{\lambda}{6}\cdot r_0^2 \Delta z$$

and for $\Lambda(z) = \left|\ln\left(\frac{r(\theta_T, z)}{r(\theta_0, z)}\right)\right| = \left|\ln\left(\frac{r_T}{r_0}\right)\right|$. Under these calculations, substituting (4.5) and letting $\lambda$ take its extreme value $\lambda = \frac{\Delta z}{V_M}$, the volume becomes $V = V_M$ and the bounds (4.12) for $\mu$ are written as

$$\left|\left(\frac{\mu}{\Delta\theta\Delta z}\right)_{\lambda=\frac{\Delta z}{V_M}}\right| \leq \frac{1}{\Delta\theta}\left|\ln\left(\frac{r_T}{r_0}\right)\right| + \left|\frac{\Delta\theta}{2} + \frac{\Delta z}{6V_M} \cdot \left[-r_0^2 + \frac{1}{\Delta\theta\Delta z}\int_{\theta_0}^{\theta_T}\left(T \cdot r(\theta,T)^2 - S \cdot r(\theta,S)^2\right)d\theta\right]\right|$$

thus obtaining mean angle's inequality consisting *Criterion 3*.                              QED

### 3.5 A fourth criterion, another necessary condition for matching: Associating the length of the contact curves (fixed and rotating chains) with the maximum allowed geometric diversification of fragments' upper surface

In this subsection, we will state a fourth criterion which, once more, is a necessary condition for actual matching of the two fragments A, B. The content of this criterion can be intuitively described as follows: Suppose that the two fragments are in contact at the *k*-th pixel of $C^F$ and the M-th pixel $C^R$; at this position, the starting $B^S_k$ and terminal $B^T_k$ barrier surfaces are unambiguously defined. In the process of testing if this position is an actual matching one, fragment B is rotated around an axis vertical to the *xy*-plane passing through the contact point of the two fragments that we momentarily consider it fixed. As fragment B is rotated, in many cases it may not intersect the terminal barrier plane or it may intersect it so as to form a rotating escarpment $E^R_{k,m}$ too wide and too "distant" from the fixed escarpment. Hence, the question arises "how to associate the maximum allowed width of $E^R_{k,m}$ with the maximum acceptable volume $\tau^T$ of $V_{k,m}$?". Under certain very plausible conditions, that seem to hold in practice, one may express the width of $E^R_{k,m}$ by means of the length of the rotating chain $\Gamma^R_{k,m}$, given that the common bottom plane remains the same. In other words, the aforementioned question is now rephrased to "how long can the rotating chain $\Gamma^R_{k,m}$ be, given that matching *Criteria 1* and *3* are satisfied?". An answer to this question will be given in the analysis that follows.

**Proposition 1**

Suppose that in a plane there are a curve $\Gamma^F$ we call fixed, another curve $\Gamma^R$ we call rotating and two parallel barrier straight line segments, an initial one $\varepsilon^I$ and a terminal one $\varepsilon^T$. Suppose, moreover, that the area of the domain bounded by this four curves $\Gamma^F$, $\Gamma^R$, $\varepsilon^I$, $\varepsilon^T$ is kept fixed, equal to *a* and that the integral of angles $\int_{T_A}^{T_B}\arctan\left(\frac{\dot{y}(t)}{\dot{x}(t)}\right)dt$

is also known and equal to $\gamma$. Then if curve $\Gamma^F$ is given and it is fixed, the length of the variable curve $\Gamma^R$ cannot exceed the value $\dfrac{a}{L_C/2}\sqrt{\dfrac{1}{\tan\left(\dfrac{\gamma}{T_B-T_A}\right)^2}+1}$.

*Proof* - Given that the length of a Jordan curve with independent variable $t \in [T_A, T_B]$ is $L_C = \int_{T_A}^{T_B}\sqrt{\dot{x}(t)^2 + \dot{y}(t)^2}\,dt$, while the area confined by such a closed curve is

$$E_C = \int_{T_A}^{T_B}(x\dot{y} - y\dot{x})dt,$$ we employ the error function

$$f(x,\dot{x},y,\dot{y}) = \sqrt{\dot{x}^2 + \dot{y}^2} - k\cdot\arctan\left(\frac{\dot{y}}{\dot{x}}\right) - \lambda(x\dot{y} - y\dot{x}).$$

Then the problem can be stated as follows: maximize quantity $J = \int_C f(x,\dot{x},y,\dot{y})dt$ under the assumptions that $\Gamma^F$, $\varepsilon^I$ and $\varepsilon^T$ are known, as well as that equations $a = \int_C (x\dot{y} - y\dot{x})dt$ and $\gamma = \int_C \arctan\left(\dfrac{\dot{y}(t)}{\dot{x}(t)}\right)dt$, hold. Forming the Euler-Lagrange equations we obtain

$$-\lambda\begin{bmatrix}\dot{y}\\-\dot{x}\end{bmatrix} = \frac{d}{dt}\begin{bmatrix}\lambda y + k\dfrac{\dot{y}}{\dot{x}^2+\dot{y}^2} + \dfrac{\dot{x}}{\sqrt{\dot{x}^2+\dot{y}^2}}\\ -\lambda x - k\dfrac{\dot{x}}{\dot{x}^2+\dot{y}^2} + \dfrac{\dot{y}}{\sqrt{\dot{x}^2+\dot{y}^2}}\end{bmatrix}.$$

Now for compactness we define the auxiliary vectors $\vec{c}(t) = (x(t), y(t))$, $\vec{r}(t) = (\dot{x}(t), \dot{y}(t))$ and $\vec{n}(t) = (\dot{y}(t), -\dot{x}(t))$, the corresponding unit vectors $\hat{r}(t)$ and $\hat{n}(t)$ and the corresponding norms $r(t) = |\vec{r}(t)|$ and $n(t) = |\vec{n}(t)|$. After some straightforward calculations we obtain the differential equation below

$$2\lambda \cdot \vec{r} = \frac{d}{dt}\left[-k\frac{\hat{r}}{r} + \hat{n}\right] \quad (5.1)$$

By integrating we obtain

$$2\lambda \cdot (\vec{c} - \vec{c}_A) = -k\frac{\hat{r}}{r} + \hat{n} - \left(-k\frac{\hat{r}_A}{r_A} + \hat{n}_A\right) \quad (5.2),$$

where $\vec{c}_A = \vec{c}(T_A)$, $\hat{r}_A = \hat{r}(T_A)$, $r_A = r(T_A)$, $\hat{n}_A = \hat{n}(T_A)$.

We perform the dot product of both sides of this equation with $\vec{r}$ and we solve the resulting differential equation to obtain

$$\left| \vec{c}(t) - \vec{c}_A + \frac{1}{2\lambda}\left(-k\frac{\hat{r}_A}{r_A} + \hat{n}_A\right)\right|^2 = -\frac{k}{\lambda}(t - T_A) + \frac{1}{4\lambda^2}\left|-k\frac{\hat{r}_A}{r_A} + \hat{n}_A\right|^2 \quad (5.3).$$

We define $\vec{K}_A = \vec{c}_A - \frac{1}{2\lambda}\left(-k\frac{\hat{r}_A}{r_A} + \hat{n}_A\right)$.

We also perform the dot product of (5.2) with $\vec{n}$ and we integrate the resulting differential equation to obtain

$$E_C = \frac{1}{2\lambda} \cdot L_R + (\tilde{c}_\Gamma - \tilde{c}_A) \cdot \vec{K}_A \quad (5.4),$$

where $\tilde{c}_\Gamma \perp \vec{c}_\Gamma$ and $\tilde{c}_A \perp \vec{c}_A$. The length of the curve that satisfies both (5.2) and (5.3) is calculated

$$L_R = \int_{T_A}^{T_B} r(t)dt = -\frac{k}{2\lambda}\cdot\left[\sqrt{\frac{1}{r_A^2} - 4\frac{\lambda}{k}(T_B - T_A)} - \frac{1}{r_A}\right] \quad (5.5).$$

We let $a(t) = \arctan\left(\frac{\dot{y}}{\dot{x}}\right)$ and we employ (5.1) in order to compute the total angle of the vectors tangent to the curve $C$, while we also employ the relations $\vec{r}(t) = (r(t)\cos(a(t)), r(t)\sin(a(t)))$ and $\vec{n}(t) = (r(t)\sin(a(t)), -r(t)\cos(a(t)))$. After performing the dot product with $\vec{r}$ and $\vec{n}$ separately with (5.1), we obtain that $\dot{a}(t) = 0$. In addition the boundary conditions for the Euler-Lagrange equations are

$$f\big|_{t=T_A} = r_A - k\cdot\arctan(a(T_A)) - \lambda\vec{n}_A\cdot\vec{c}_A = -\frac{\lambda E_C}{T_\Gamma - T_A} - \frac{k\mu_C}{T_\Gamma - T_A}$$

$$\frac{\partial}{\partial \vec{r}} f\bigg|_{t=T_B} = \lambda\vec{c}_\Gamma + k\frac{\hat{r}_\Gamma}{r_\Gamma} - \hat{n}_\Gamma = 0.$$

Combining the above conditions with (5.4) and (5.5) we compute the desired result for the maximum allowed length $L_R = \frac{E_C}{|\vec{c}_A|}\sqrt{\frac{1}{\tan\left(\frac{\gamma}{T_\Gamma - T_A}\right)^2} + 1}$ QED

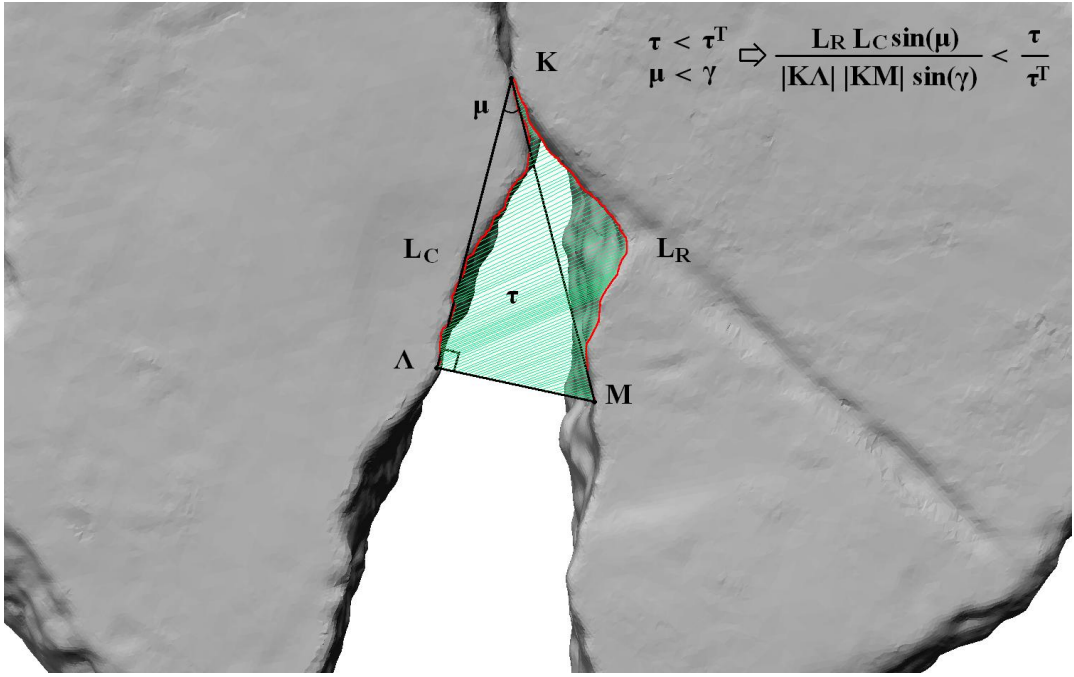

**Fig. 6:** Visualization of Criterion 4 to an arbitrary relative placement of the two fragments. The maximum acceptable length $L_R$ of $\Gamma^R$ is computed via the area $\tau$ enclosed by $E^F$, $E^R$ and $M$, $\Lambda$ the fixed length of comparison $L_C$ and the integral of angles $\mu$ over $\Gamma^F$ and $\Gamma^R$.

To get an idea, we let the length of the fixed curve $\Gamma^F$ be $L^C = 100$ pixels, the maximum allowed area enclosed by curves $\Gamma^F$, $\Gamma^R$, $\varepsilon^I$, $\varepsilon^T$ be $\tau^T = 1000$ pixels and the mean angle of the vectors tangent to the previous curve be $\bar{\gamma} = 0.1745$ rad or equivalently $\bar{\gamma} = 10°$. Then the maximum allowed length $L^{EX}$ of the rotating chain is $L_R \approx 115$ pixels. As already mentioned in Sect. 3.1 we use the term pixels to denote the points of the digital contour of fragment's upper plane surface.

In order to apply the aforementioned *Criterion 4*, it is necessary to give an estimation of the upper-bound $a$ for the area enclosed by $\Gamma^F$, $\Gamma^R$, $\varepsilon^I$, $\varepsilon^T$ as referred in Lemma. This upper-bound may either be taken ad-hoc or it may be estimated as follows: It is quite logical to assume that the area of all cross-sections of the domain $V_{k,m}$ follows a normal distribution. We have applied Kolmogorov-Test to check this hypothesis (level of significance=0.001) and the obtained results did not contradict this hypothesis at all. Therefore a reasonable upper-bound $\alpha$ for the considered area is $a = 3.1 \dfrac{\tau^T}{\Delta z}$, since 99.9% of the normal population of the cross sections remain less than this bound; as always $\tau^T$ is the maximum acceptable volume of $V_{k,m}$ and $\Delta z$ the z-difference between the upper and bottom planes.

# 4 Demonstration of the Way Matching Criteria Assemble to a Consistent Hierarchy

In order to develop an algebraic interpretation and justification of the structure of the algorithm, we will employ an abstraction of the elements of the matching procedure by defining $D^n$ to represent simply connected $n$-dimensional open domains embedded in $\mathbf{R}^N$ for a constant $N$, $n \leq N$, and morphisms $t: D^{n-1} \times D^{n-1} \rightarrow D^n$ to represent formation of an $n$-dimensional bounded domain from 2 open $(n-1)$–dimensional boundary elements. By adding morphisms $m$ from domains $D^n$ to a class of their measures $\mu^n$ we form a category $\mathbf{A}^n$ with objects $D^{n-1} \times D^{n-1}, D^n, \mu^n$ and morphisms $t, m$ and the binary relations of sets that domains carry from the category of relations **Rel**. Two different pairs in $D^{n-1} \times D^{n-1}$ map through a morphism in $t$ to two different domains in $D^n$, implying that $t$ is a class of monomorphisms, usually denoted by the arrow $\hookrightarrow$. On the other hand, to obtain two different images of the same $n$-domain in $\mu^n$, we need two different maps in the class of $m$, implying that $m$ is a class of epimorphisms, usually denoted by the arrow $\twoheadrightarrow$. Thus, elements of $\mathbf{A}^n$ respect the sequence of morphisms $D^{n-1} \times D^{n-1} \hookrightarrow D^n \twoheadrightarrow \mu^n$, which is well known to be exact if the kernels of the morphisms exist, namely if zero morphisms exist. By defining a zero element for $\mathbf{A}^n$, kernels and co-kernels of $t$ and $m$ for this sequence of morphisms are sure to exist. In Appendix 2, by attaching to $\mathbf{A}^n$ an auxiliary binary operator, it is demonstrated that $\mathbf{A}^n$ has $\emptyset$ as its zero object. Thus kernels and co-kernels of $\mathbf{A}^n$ respect the following short exact sequence

$$\emptyset \rightarrow D^{n-1} \times D^{n-1} \xrightarrow{t} D^n \xrightarrow{m} \mu^n \rightarrow \emptyset \quad (6.1)$$

The fact that (6.1) is exact makes $\mathbf{A}^n$ a normal category as $\ker m = \operatorname{im} t$ and $\operatorname{coker} t = \operatorname{im} m$. In Appendix 2 it is also given a definition of binary products (Cartesian products of elements with morphisms $t$ and $m$ applied point-wise) and a definition for binary co-products, making $\mathbf{A}^n$ an abelian category. Thus, collecting all $\mathbf{A}^n$ to a larger category $\mathbf{A}$ and defining boundary operators from the objects of $\mathbf{A}^n$ to the objects of $\mathbf{A}^{n-1}$ we can reformulate (6.1) as a short exact sequence of chain complexes.

Since $D^n$ are bounded domains, they accept a boundary operator $\partial$ that maps each domain to its boundary. In order to render $\partial D^n$ an element of $D^{n-1}$ we need to remove a fixed point $p$ from $\partial D^n$ thus writing boundary operator in the form $\partial_p^n$. Concerning elements of $\mu^n$, a boundary operator can be defined in two steps: 1) We consider a differentiable monomorphism $\varphi$ in the class of $t$, whose total differential is defined as pushforward $\varphi_*$ between tangent spaces of $D^{n-1}$ and $D^n$. 2) As $\mu^n$ has elements in the quotient space $(D^n \to \mathbb{R})/\operatorname{im} m$, we consider a differentiable function $f : D^n \to \mathbb{R}$ and a measure $\mu \in \operatorname{im} m$ and we combine them to define $I \in \mu^n : (D, f, \mu) \mapsto \int_D f d\mu$ and we evaluate $1^{\text{st}}$ order variation of $I$ w.r.t. $\varphi$. By the short exact sequence (6.1) it follows that $\mu^n$ is isomorphic to $D^n/\operatorname{im}(t)$ and, since $I$ is defined on measures accepting infinitesimals, we can reformulate $I$ by letting $D \in D^n/\operatorname{im}(\varphi)$. Moreover, since $I$ is in the quotient of $\operatorname{im} m$, it asks for $D$ to be in the $\ker m = \operatorname{im}(\varphi)$. Thus

$$I = \int_D f \circ \varphi \, d\mu = \int_D f \circ \varphi \, \Theta_\mu(\varphi) \, \omega = \int_{\varphi^{-1}(D)} f \circ \varphi \, \Theta_\mu(\varphi) \, \varphi^* \omega$$

Where $\omega$ denotes the volume form in $D^n$, $\varphi^*\omega$ is the n-1 differential form in $\varphi^{-1}(D)$ offered by the pull-back of $\omega$ by $\varphi$ and $\Theta_\mu(\varphi)$ is the density of the measure $\mu$ w.r.t. the volume of $D$. This density is guaranteed to exist by the Radon-Nikodym theorem and due to the fact that $\omega = 0 \Rightarrow \operatorname{vol}(D) = 0 \Rightarrow D = \emptyset \Rightarrow \mu(D) = 0$ satisfying absolute continuity condition. Moreover, the pre-image $\varphi^{-1}(D)$ of $D$ in $D^{n-1} \times D^{n-1}$ exists and it is unique for any $D \in \operatorname{im}(\varphi)$ since $\varphi$ is a monomorphism.

The functional variation of $\varphi$ can be obtained, at least infinitesimally, by the action of a vector flow in $D^n$, $\rho_{\eta_* X}^\epsilon(\varphi) = (\exp(\epsilon \eta_* X))[\varphi]$, with $\eta$ an arbitrary differentiable monomorphism in the class of $t$, $X$ an arbitrary vector field defined in the tangent space of $\varphi^{-1}(D)$ and exp the exponential map. Then, first order variation of $I$ under functional variation of $\varphi$ reads

$$\delta I = \frac{d}{d\epsilon} \int_{\varphi^{-1}(D)} (f \, \Theta_\mu) \circ \rho_{\eta_* X}^\epsilon(\varphi) \, \rho_{\eta_* X}^\epsilon(\varphi)^* \omega \bigg|_{\epsilon=0} \qquad (6.2)$$

But both $\frac{d}{d\epsilon}(f \, \Theta_\mu) \circ \rho_{\eta_* X}^\epsilon(\varphi)\big|_{\epsilon=0}$ and $\frac{d}{d\epsilon} \rho_{\eta_* X}^\epsilon(\varphi)^* \omega\big|_{\epsilon=0}$ are Lie derivatives acting on a function and a differential form respectively. Thus $\frac{d}{d\epsilon}(f \, \Theta_\mu) \circ \rho_{\eta_* X}^\epsilon(\varphi)\big|_{\epsilon=0} = \eta_* X(f \, \Theta_\mu)$ and by Cartan's identity and the fact that $\omega$ is exact we have $\frac{d}{d\epsilon} \rho_{\eta_* X}^\epsilon(\varphi)^* \omega\big|_{\epsilon=0} = (\mathbf{d}\iota_{\eta_* X} + \iota_{\eta_* X}\mathbf{d})\omega = \mathbf{d}\iota_{\eta_* X}\omega$, where $\iota_{\eta_* X}$ denotes the contraction (interior product) of a differential form with $\eta_*X$ and $\mathbf{d}$ denotes the exterior derivative. Hence, by substituting in (6.2) we obtain

$$\delta I = \int_{\varphi^{-1}(D)} \eta_*X(f\Theta_\mu)\, \varphi^*\omega + \int_D f\Theta_\mu\, \mathbf{d}\iota_{\eta_*X}\omega$$

Using the expansion $\mathbf{d}(f\Theta_\mu\, \iota_{\eta_*X}\omega) = f\Theta_\mu\, \mathbf{d}\iota_{\eta_*X}\omega + \eta_*X(f\Theta_\mu)\omega$ and applying the Stokes theorem we have $\int_D f\Theta_\mu\, \mathbf{d}\iota_{\eta_*X}\omega = \int_{\partial D} f\Theta_\mu\, \iota_{\eta_*X}\omega - \int_D \eta_*X(f\Theta_\mu)\omega$. Thus $\delta I$ is reduced to

$$\delta I = \int_{\partial D} f\Theta_\mu\, \iota_{\eta_*X}\omega$$

Therefore, $\delta$ defines a boundary operator in $\mu^n$.

After completing the definition of boundary operators for all elements of the short exact sequence (6.1), we can distribute (6.1) in chain complexes $\mathcal{E} = (D^{n-1} \times D^{n-1}, \partial^{n-1})$, $\mathcal{D} = (D^n, \partial^n)$, $\mathcal{M} = (\mu^n, \delta^n)$, as the following commutative diagram shows

$$\begin{array}{ccccc}
\downarrow \partial_p^n & & \downarrow \partial_p^{n+1} & & \delta^{n+1} \downarrow \\
D^{n-1} \times D^{n-1} & \xrightarrow{t^n} & D^n & \xrightarrow{m^n} & \mu^n \\
\downarrow \partial_p^{n-1} & & \downarrow \partial_p^n & & \delta^n \downarrow \\
D^{n-2} \times D^{n-2} & \xrightarrow{t^{n-1}} & D^{n-1} & \xrightarrow{m^{n-1}} & \mu^{n-1} \\
\downarrow \partial_p^{n-2} & & \downarrow \partial_p^{n-1} & & \delta^{n-1} \downarrow \\
\vdots & \xrightarrow{t^{n-2}} & \vdots & \xrightarrow{m^{n-2}} & \vdots
\end{array} \qquad (6.3)$$

Then by "zig-zag lemma" there exist boundary maps $\tilde{\delta}^n$ sending the homology groups $H_n(\mathcal{M})$ to the homology groups $H_{n-1}(\mathcal{E})$ making the following sequence exact

$$\cdots \xrightarrow{\tilde{\delta}^{n+1}} H_n(\mathcal{E}) \xrightarrow{t^n} H_n(\mathcal{D}) \xrightarrow{m^n} H_n(\mathcal{M}) \xrightarrow{\tilde{\delta}^n} H_{n-1}(\mathcal{E}) \xrightarrow{t^{n-1}} \cdots$$

By their definition the homology groups read $H_n(\mathcal{M}) = \ker \delta^n / \mathrm{im}\, \delta^{n+1}$ and $H_{n-1}(\mathcal{E}) = \ker \partial_p^{n-2} / \mathrm{im}\, \partial_p^{n-1}$. Since, by its definition, $\delta$ is linear, $H_n(\mathcal{M})$ can be written in the form of a map $L_{(a,b)} : (\tilde{I}^n, \delta^{n+1}I^{n+1}) \mapsto a\tilde{I}^n + b\delta^{n+1}I^{n+1}$, where $\tilde{I}^n \in \mu^n : \delta^n \tilde{I}^n = 0$ and $I^{n+1} \in \mu^{n+1}$. On the other hand, given an element $\tilde{D}_n \in D^n$ : $\delta^n I^n\left(m^n(\tilde{D}_n)\right) = 0$, the commutative diagram (6.3) implies that $m^{n-1}(\partial_p^n \tilde{D}_n) = 0 \Rightarrow \partial_p^n \tilde{D}_n \in \ker m^{n-1} \Leftrightarrow \partial_p^n \tilde{D}_n \in \mathrm{im}\, t^{n-1}$. Therefore for each such $\tilde{D}^n$ there is a unique pair $(\tilde{D}_1^{n-2}, \tilde{D}_2^{n-2})$. By the zig-zag lemma, this pair is offered by $\tilde{\delta}^n$ and belongs to $H_{n-1}(\mathcal{E})$, namely it is a pair of cycles in $D^{n-2}$ over a corresponding pair of boundaries in $D^{n-1}$, which are joint in the fixed point $p$. Summarizing the aforementioned, bounded domains formed by 2 boundary elements so as to correspond to stationary points of a Lagrangian integral $\tilde{I}^n$ with restrictions described by $\delta^{n+1}I^{n+1}$, have orbits in the $C^{n-2}$ cycles over the boundaries of the generating boundary elements.

Stating the aforementioned analysis in $\mathbf{R}^3$, for the problem in hand, the following proposition results

**Proposition 2**

Consider two surface patches $E_F$, $E_R$, that are boundary elements of a bounded domain $D$ in $\mathbf{R}^3$, which renders a Lagrangian $I^3$ stationary.

A) Formation of D by $E_F$, $E_R$ has orbits in $C^2$ cycles over the permutations of the points of $\partial E_R$ that are joint with a fixed point $p$ on $\partial E_F$ and vice-versa.

Consider moreover pairs $(\Gamma_S^F, \Gamma_T^F)$ and $(\Gamma_S^R, \Gamma_T^R)$, where $(\Gamma_S^F, \Gamma_T^F)$ are contour parts of $\partial E_F$ and $(\Gamma_S^R, \Gamma_T^R)$ are contour parts of $\partial E_R$, defined in the quotient space of each contact point selection in $\partial E_R$. Suppose that there is a surface patch $E_R'$ with $\Gamma_S^R, \Gamma_T^R$ lying on its boundary, which renders Lagrangian $I^2 + \lambda_2 \delta I^3$ stationary.

B) Then, formation of $E_R'$ by $(\Gamma_S^R, \Gamma_T^R)$ has orbits in $C^1$ cycles over the binary relation $(\partial_p \Gamma_S^R, \partial_p \Gamma_T^R) \sim (\partial_p \Gamma_S^F, \partial_p \Gamma_T^F)$, namely over the joint permutations of the ending points of $(\Gamma_S^R, \Gamma_T^R)$ and $(\Gamma_S^F, \Gamma_T^F)$.

C) These ending points are fully determined as stationary points of a Lagrangian $I^1 + \lambda_1 \delta I^2$.

The above sequence of stationary points – orbits – stationary points – … is exact implying that there is no global Lagrangian formulation in the Error – Restrictions setting for the simultaneous extremization of the domains and their boundary elements.

We would like to emphasize that the sequence described in Proposition 2 is exactly the hierarchy of the criteria of the introduced methodology. In other words, it has been proved that the employed hierarchy of the criteria is necessary to ensure that boundary elements that do not belong to orbits of the stationary points of a Lagrangian, will be excluded a priori.

# 5 Application of the Method

## 5.1 The employed method for 3D scanning of wall-painting fragments

We have used a prototype dedicated system of IMETRIC GMBH, specially configured for archaeological artifacts. The system consists of 2 cameras and a high quality DLP projector, with nominal scanning precision in vivo of 3-7μm. The items to-be-scanned were placed on a reference plate bearing photogrammetric targets. The reference plate has been also scanned and measured empty and the *xyz*-coordinates of each one of its targets were obtained. The average accuracy of this measurement was 5μm/1m. The reference plate was placed on a rotation device and for each angle of rotation the coordinates of the reference targets were measured. Next, the fragments were placed one by one on the reference plate and the coordinates of the sampling points on their surface were measured on the basis of the reference plate measurements. The surface points' coordinates have been obtained by photogrammetric reconstruction from two different 2D projections using the methodology of structured-light scanning as it is extensively described in [20]. The scanning process was performed with sampling resolution of 0.14mm.

## 5.2 Description of the process applied for automated reassembly of fragmented objects on the basis of the aforementioned criteria

Suppose that we have $N$ available fragments and their 3D representations obtained as described in Sect. 5.1. In order to achieve an optimal reassembly of these fragments, we have applied the process consisting of the steps described below

*Step 1 – Choice of the proper parameters*

First we set an angular step $\delta\theta=1^o$ to generate all rotated versions of each fragment. Next, we choose the length of contact curve on fixed fragments' upper surface, $L_C$, to be a percentage, here 15%, of the mean length/perimeter of all fragments' upper surface boundary curves. We, also, compute the mean area of all fragments' lateral surface and use it, together with $L_C$, in order to define via inequality of Sect. 3.4 the volume threshold, $\tau^T$, so that a maximum average gap of $h$mm between actually matching fragments is acceptable. We start from a very small value of $h$, say $h=0.4$mm, to account for almost perfect matching between adjacent fragments. Then we sequentially increase $h$ up to 1.2mm to allow for larger gaps between matching fragments due to more serious wear. In addition, for each value of $h$ separately, we set the threshold area that the contact curves enclose, $E_R=hL_C$, together with the threshold angular deviation $\gamma = \arctan\left(\dfrac{h}{L_C}\right)$ for computing (Sect. 3.5).

*Step 2 – Application of the Criteria considering the larger fragment as the fixed one*

We spot the fragment with the larger upper surface area, say $F_1$, and we let it be the fixed fragment of the matching process. Subsequently, we look for possible matching positions between $F_1$ and all other fragments, which are considered to be the rotating ones, according to the analysis introduced in *Section 3* and by application of the developed criteria in the following order:

A) First, we apply *Criterion 4* checking the relative lengths of the contact curves in the common upper surfaces of the fragments (Sect. 3.5).

B) Second, if *Criterion 4* is satisfied, we examine if "contact surfaces" of the considered two fragments at the specific relative position satisfy *Criterion 3* concerning their geometric similarity (Sect. 3.4).

C) Third, in case that *Criterion 3* is also satisfied, *Criterion 2* checks possible overlapping between the two fragments at the position in hand, both locally and overall, acting as described in Sect. 3.3.

D) If the two considered fragments, at the specific relative position, pass these 3 rejection filters, then and only then the system proceeds in testing the final *Criterion 1* checking if the volume of the gap between the two fragments is smaller than the properly predefined threshold $\tau^T$. If *Criterion 1* is, also, satisfied at the relative fragments position, then the system characterizes the specific relative placement of the two fragments as a matching one.

*Step 3 – Merging of matching fragments to generate an island*

If application of *Step 2* above offers matching of fragment $F_1$ with a number of other fragments then we virtually merge these matched fragments to form an island $I_1$.

Subsequently, we let $I_1$ play the role of $F_1$ and we repeat *Steps 2* and *3* until no further matching is reported.

*Step 4 – Repeating the reassembly process for the non-matched fragments*

By the end of *Step 3* it is probable that there are fragments not belonging to island $I_1$. Among them we spot the fragment with the larger upper surface area, we let it play the role of $F_1$ and we repeat *Steps 2, 3* and *4* thus obtaining a set of islands $I_n$. The process ends when the non-matched fragments have been exhausted and no further matches are reported.

*Step 5 – Repeating the process for a new value of constant "h"*

By the end of *Step 4* we have ended up with a number of islands $I_n$, some of which may be single fragments. At this point, we increase the value of *h* by a small quantity, say 0.2mm, in order to allow for a larger gap between two adjacent fragments and we repeat *Steps 1-5*.

In case that, during this process, false positive/inconsistent matches had been generated, then one should turn back to the point that inconsistency had appeared for the first time and one should check all sequences of possible matches to find the one that lifts the inconsistency. However, we would like to stress that, in our real data, application of the process described in Sect. 5.3, no false positive matches have been reported during the matching process. If one considers the fact that we have applied the process to prehistoric wall-paintings excavated near the end of the 19th century, where it is reasonable to expect that these fragments have suffered serious wear, then one may claim that the correctness and uniqueness constitute a major merit of the methodology introduced here.

**5.3 Application of the matching process to the virtual reconstruction of fragmented prehistoric wall-paintings**

First, the authors have tested the method and the system in the case of two objects that have been broken on purpose. The reason for applying the methodology and the related system in this artificial case has to do with the difficulty of finding matching pairs of fragments in the actual case of Tyrins wall-paintings. In fact, due to the 3.300 years that they have passed from the moment the wall paintings were fragmented, it is logical to expect that many fragments will be missing, as well as that all fragments will suffer from considerable wear. In addition, one does not a priori know the thematic content of the painting and therefore one does not know the exact solution of the fragmented object that must be reassembled. For all these reasons, first, a flagstone was placed inside a sealed sack, which in turn was thrown from a certain height. As a result the flagstone was broken into 9 fragments. We have repeated the same process with a second flagstone which was broken in 15 pieces. The ensemble of 24 fragments have been scanned and pre-processed as described in Sections 2. Then the matching process outlined in 5.2 offered two islands, which precisely corresponded to the two fragmented flagstones.

However, the first major goal of the methodology presented here was to contribute to the reassembly of wall painting fragments of great archaeological importance. Indeed, the fresco fragments used in this study, scanned by the authors, are housed in the Prehistoric Collection of the National Archaeological Museum in Athens (inv. nos 1596, 1655, 1668, 5881-3). They come from excavations at the mycenaean palace of Tiryns led by H. Schliemann in 1885-1886 (H. Schliemann, Tiryns, Leipsig 1886, pls V, VI, XI) and the German Institute in the years 1909-1910 (D. Rodenwaldt, Die Fresken des Palastes, Tiryns II, Athen 1912, pl. III, XXI). They present a variety of decorative motifs, including spirals and schematic plants, rosettes, elaborate abstract patterns and are dated in the 14th-13th centuries BC. Some of them, with a plaster thickness of 2.5cm, belong to the smaller megaron's floor (Schliemann 1886 pl. XI, Rodenwaldt 1912 pl. XXI).

The mycenaean acropolis at Tiryns is conspicuous today for its mighty Cyclopean walls that led the epic poet Homer to call it 'well-walled' in the Iliad. It is only second in importance after Mycenae, the capital of the legendary Agamemnon.

Heinrich Schliemann, the excavator of Troy and Mycenae, and the architect Wilhelm Dörpfeld, excavated the acropolis of Tyrins in 1885 and 1886. Today, the Tiryns excavations continue under the direction of the German Archaeological Institute. The citadel was the administrative, financial and religious centre for a wide region in the 14th and 13th centuries BC.

The authors scanned 41 fragments belonging to this collection and they have applied the process of Sect. 5.2 in order to look for possible matching islands among them. The system eventually offered 9 islands of matching fragments, where each island consisted of 2-4 pieces. The corresponding matching results are presented in Tables 1 and 2, where the islands $I_n$ are given in the order the system offered them together with the fragments that they consist of and the number of the remaining fragments each time island $I_n$ is formed (e.g. see Figure 7). Dedicated conservators and expert archaeologists have tested the results offered by the system. In fact, we have furnished the dedicated personnel with the images of the matching positions automatically generated by the system. The conservators and archaeologists placed the fragments of each image manually in the depicted relative position and confirmed that all proposed matches are correct according to their knowledge and experience. Figures 7-15 manifest the correspondence between the automated reassembly offered and the manual one performed by the scholars.

Concerning the time requirements of the system we note that the methodology introduced here has been applied and executed on a processor Intel Core 2 Duo 3.2 GHz, with 4 GB RAM at 1033 MHz. During the application of the reassembly process, we have met with 2 distinct cases concerning time performance:

1) When the system did not spot matching between the considered pair of fragments, the time required for this decision never exceeded 14 seconds. This quite rapid decision is due to the rejection Criteria 3, 4 described in 3.4, 3.5. Evidently, the exact time required for the decision relies on the size of the involved fragments or islands.

2) When the system found a match, it required up to 8 minutes to offer the optimal matching position. This is due to the fact that there are numerous neighbouring relative positions of the fixed and rotating fragments where all criteria are satisfied and as a consequence must compute the volume $\tau_{k,m}$ of $V_{k,m}$ and all other quantities involved in the developed criteria many times in order to decide for the optimal matching position. In any case, we stress that the whole matching process is immediately parallelizable and therefore the overall time required for the entire reassembly process can be almost linearly minimized by increasing the number of processors employed for testing pairwise fragments matching.

**Table 1** – The sequence of fragments islands spotted by the system after 27 iterations. Island $I_1$ was generated $1^{st}$, $I_3$ $2^{nd}$, etc.

|  | Spotted Islands of Fragments |  |  |  |  |  |  |  |
|---|---|---|---|---|---|---|---|---|
|  | $I_1$ | $I_3$ | $I_7$ | $I_8$ | $I_{11}$ | $I_{18}$ | $I_{20}$ | $I_{21}$ | $I_{23}$ |
| Fragments merged to each Island | $F_3$ | $F_{11}$ | $F_{20}$ | $F_4$ | $F_8$ | $F_{14}$ | $F_9$ | $F_{19}$ | $F_{24}$ |
|  | $F_{17}$ | $F_{27}$ | $F_{23}$ | $F_{33}$ | $F_{22}$ | $F_{28}$ | $F_{12}$ | $F_{40}$ | $F_{39}$ |
|  | $F_{25}$ | $F_{32}$ | - | - | - | $F_{30}$ | $F_{14}$ | - | - |
|  | $F_{31}$ | - | - | - | - | - | $F_{29}$ | - | - |
| Fragments not matched to the fixed one | 37 | 33 | 28 | 26 | 22 | 13 | 8 | 6 | 3 |

Table 1 - The islands of fragments offered by the system are presented. Island $I_1$ has been spotted $1^{st}$, $I_3$ $2^{nd}$, $I_7$ $3^{rd}$, etc. The lower subscript of $I_n$ corresponds to the number of the iteration at which the island has started to be formed. This means that island $I_2$ consists of a single fragment and it is omitted, since no matching pair of it was found during the entire matching process. The spotted islands are shown in figures 7-15, together with images of the reassembly performed by the scholars. The fragments are denoted by $F_m$ ordered as described in Sect. 5.1.

**Table 2** – Mean Euclidean Error for Each Matching Pair.

|  | Spotted Islands of Fragments | | | | | | | | |
|---|---|---|---|---|---|---|---|---|---|
|  | $I_1$ | $I_3$ | $I_7$ | $I_8$ | $I_{11}$ | $I_{18}$ | $I_{20}$ | $I_{21}$ | $I_{23}$ |
| Fragments merged to each Island | $F_3$ $F_{17}$ $F_{25}$ $F_{31}$ | $F_{11}$ $F_{27}$ $F_{32}$ - | $F_{20}$ $F_{23}$ - - | $F_4$ $F_{33}$ - - | $F_8$ $F_{22}$ - - | $F_{14}$ $F_{28}$ $F_{30}$ - | $F_9$ $F_{12}$ $F_{14}$ $F_{29}$ | $F_{19}$ $F_{40}$ - - | $F_{24}$ $F_{39}$ - - |
| Mean Euclidean distances between triangle vertices of each matching domain (mm) | - 0.510 0.499 0.350 | - 0.400 0.284 | - 0.360 | - 0.400 | - 0.398 | - 0.548 0.393 | - 0.955 0.261 0.270 | - 0.282 | - 0.444 |

Table 2 - The sequence of mean Euclidean distances is in a one to one correspondence with the fragments that are, each time, matched to the corresponding island; e.g. $F_{17}$ optimally matches $F_3$ with 0.510 mm mean Euclidean distance between the fixed and rotating escarpments, $F_{25}$ matches island $I_1$ formed by $F_3$ and $F_{17}$ with a corresponding mean error of 0.499 mm, etc. Visualisation of the related results is given in figs. 7-15.

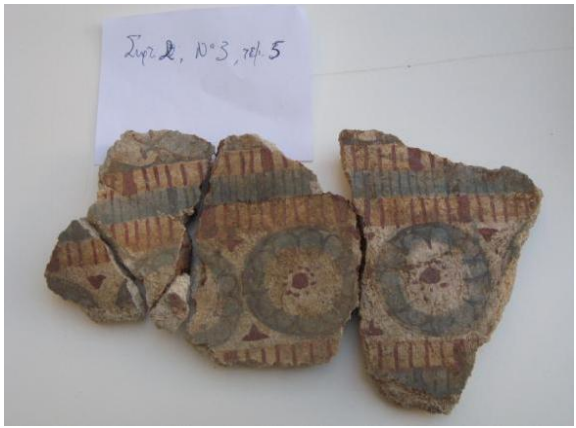
(a)

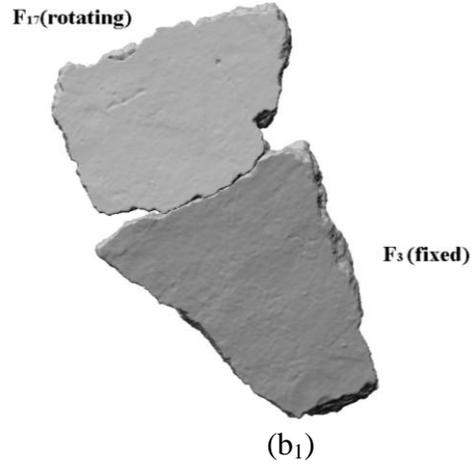
(b₁)

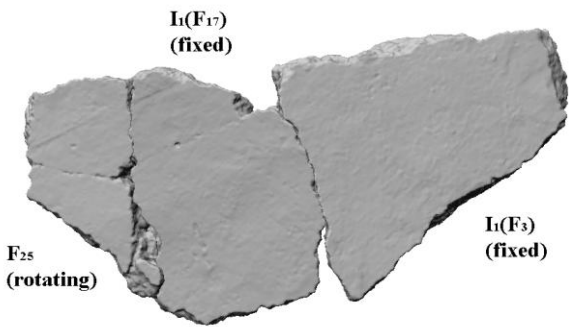
(c₁)

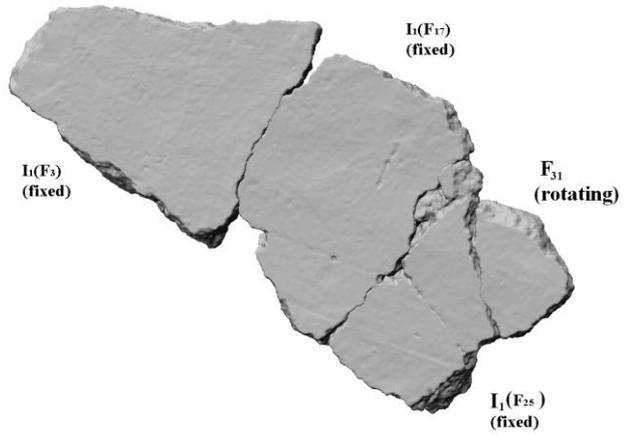
(d₁)

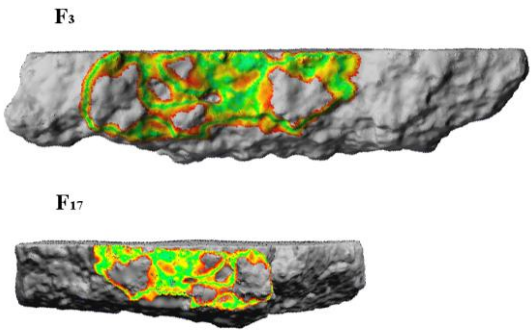
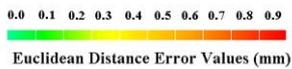
(b₂)

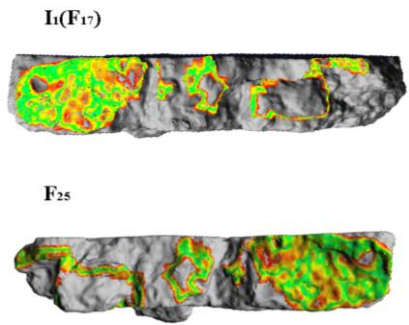
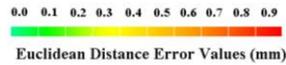
(c₂)

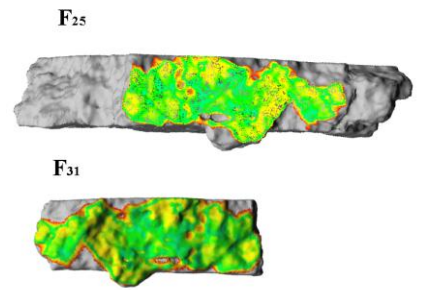
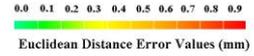
(d₂)

**Fig. 7:** Matching results and verification of island $I_1$
(a): The actual island formed by conservators according to the system suggestion.
($b_1$), ($c_1$), ($d_1$): Pair-wise matches proposed by the system and the subsequent merge of the 3D representation of the two fragments at the matching position.
($b_2$), ($c_2$), ($d_2$): Visualization of Euclidean distances between fragments surfaces at the matching positions depicted in ($b_1$), ($c_1$) and ($d_1$) respectively.

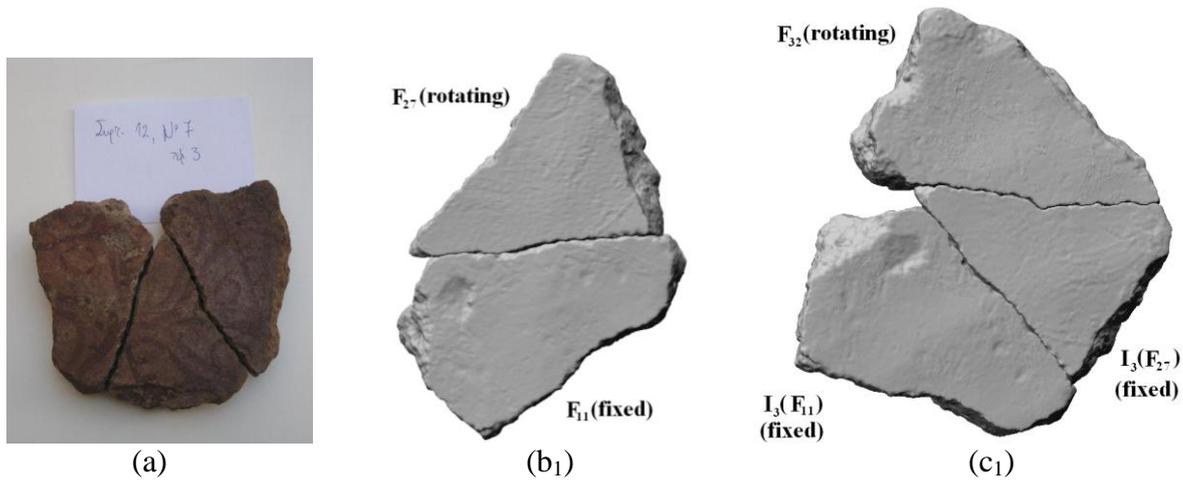

**Fig. 8:** Matching results and verification of island $I_3$. (a): The actual island formed by conservators according to the system suggestion. ($b_1$), ($c_1$): Pair-wise matches proposed by the system and the subsequent merge of the 3D representation of the two fragments at the matching position. ($b_2$), ($c_2$): Visualization of Euclidean distances between fragments surfaces at the matching positions depicted in ($b_1$) and ($c_1$) respectively.

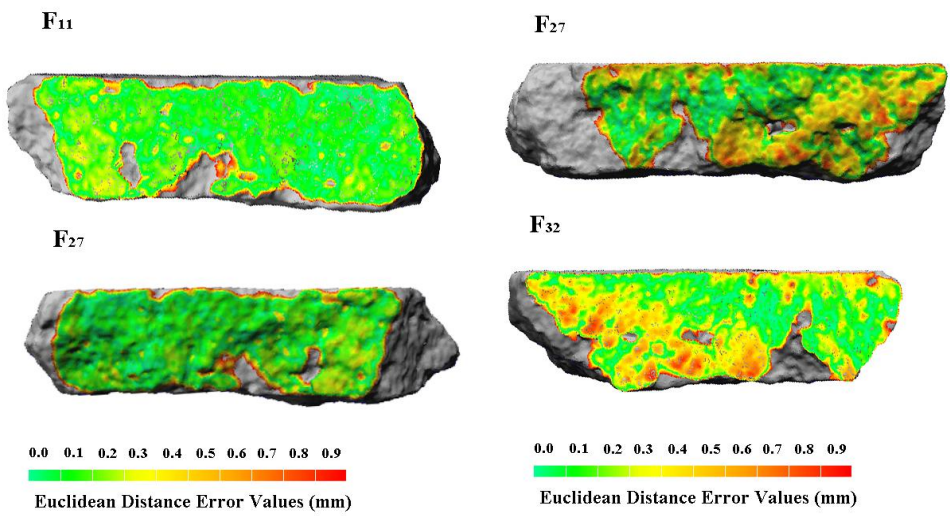

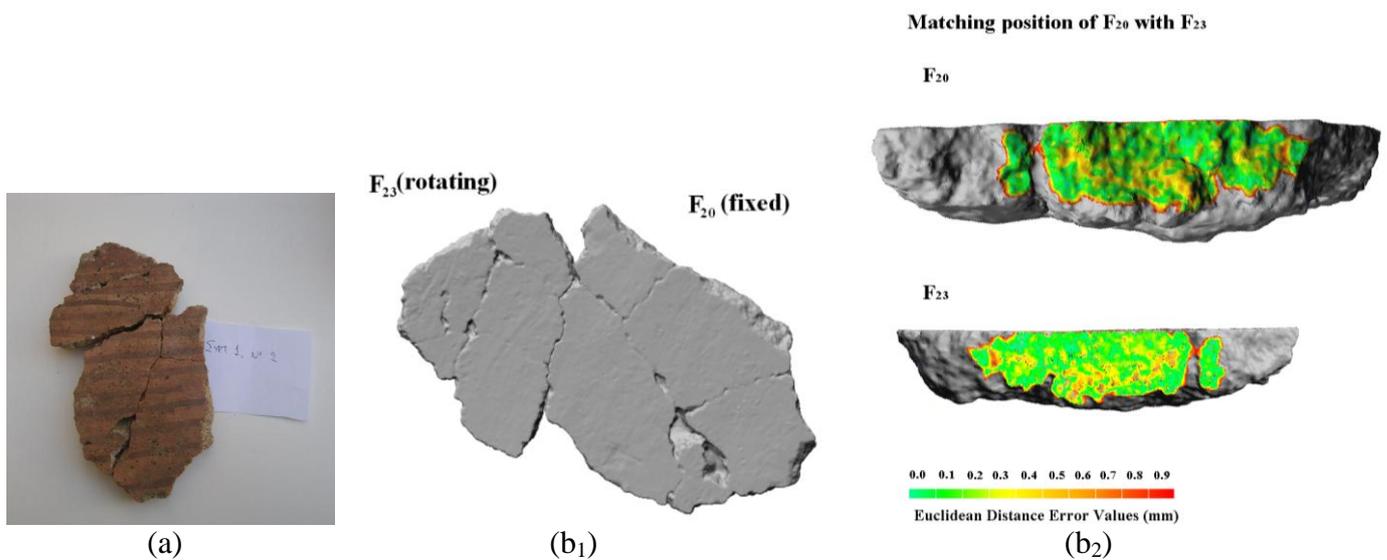

**Fig. 9:** Matching results and verification of island $I_7$. (a): The actual island formed by conservators according to the system suggestion. ($b_1$): Pair-wise matches proposed by the system and the subsequent merge of the 3D representation of the two fragments at the matching position. ($b_2$): Visualization of Euclidean distances between fragments surfaces at the matching positions depicted in ($b_1$).

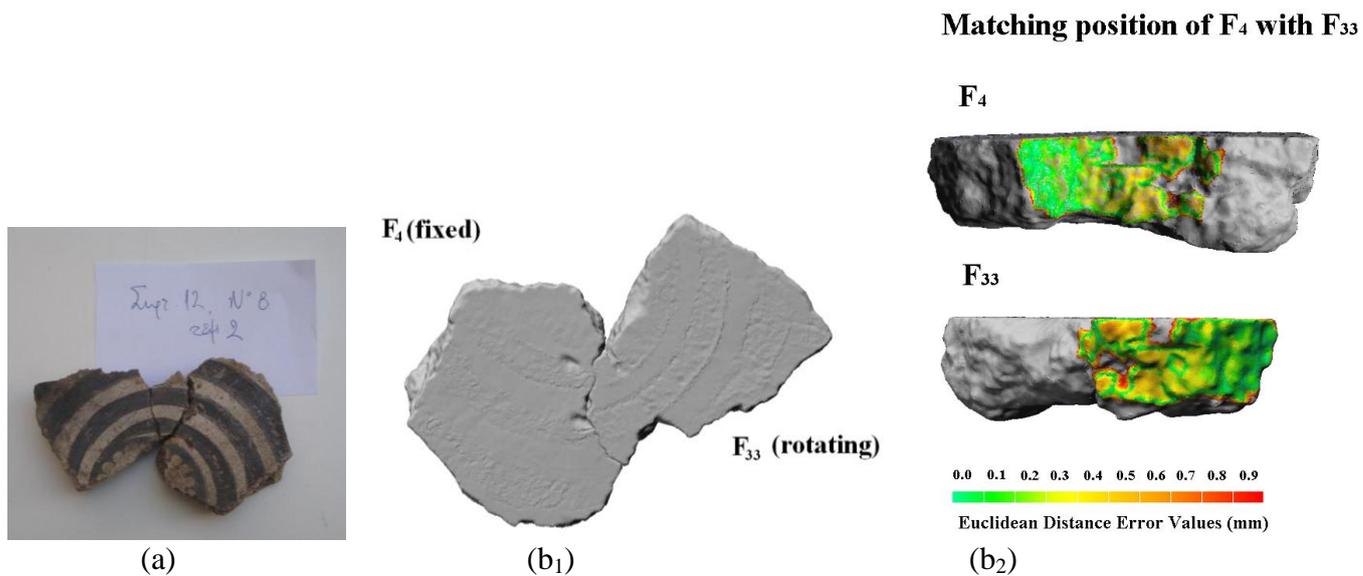

**Fig. 10:** Matching results and verification of island $I_8$. (a): The actual island formed by conservators according to the system suggestion. ($b_1$):Pair-wise matches proposed by the system and the subsequent merge of the 3D representation of the two fragments at the matching position. ($b_2$): Visualization of Euclidean distances between fragments surfaces at the matching positions depicted in ($b_1$).

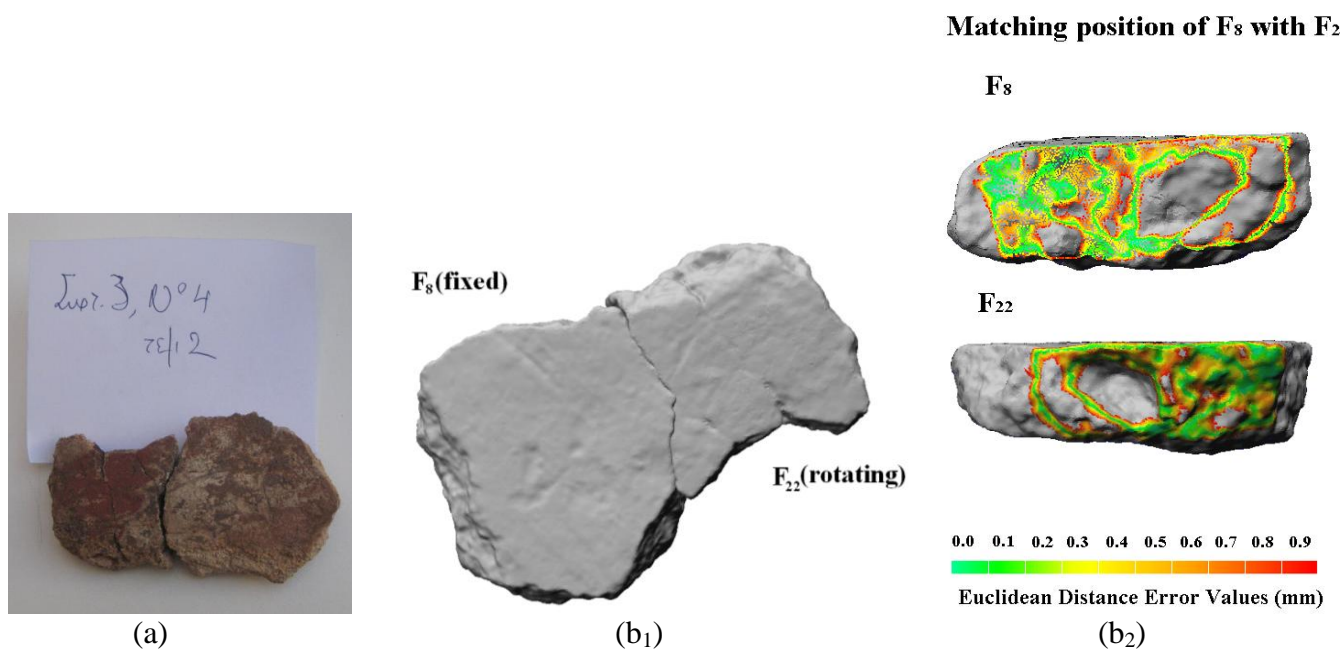

**Fig. 11:** Matching results and verification of island $I_{11}$. (a): The actual island formed by conservators according to the system suggestion. ($b_1$):Pair-wise matches proposed by the system and the subsequent merge of the 3D representation of the two fragments at the matching position. ($b_2$): Visualization of Euclidean distances between fragments surfaces at the matching positions depicted in ($b_1$).

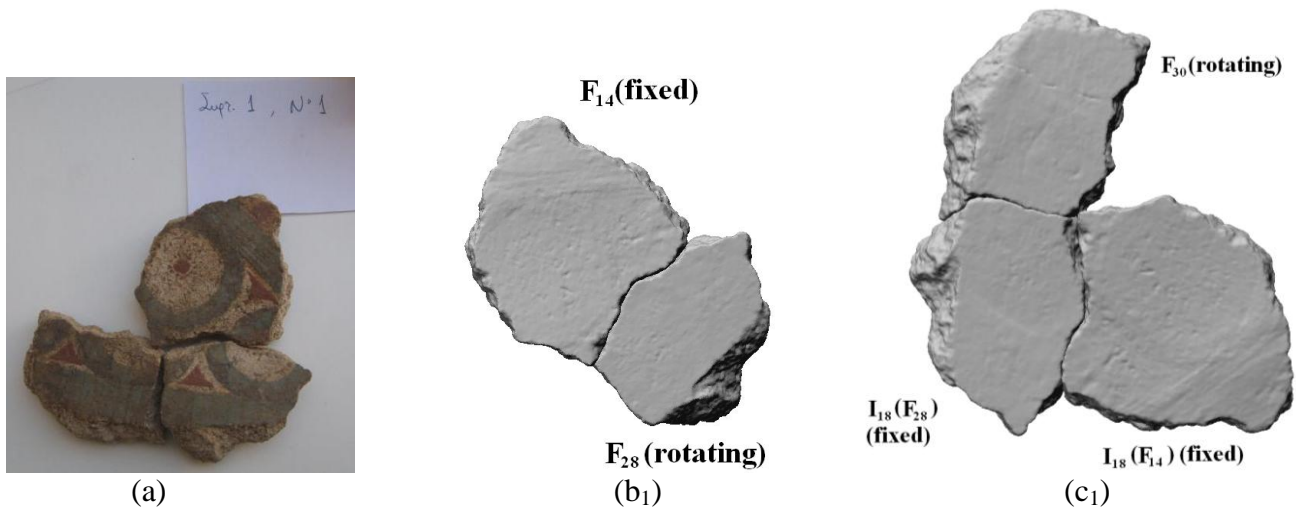

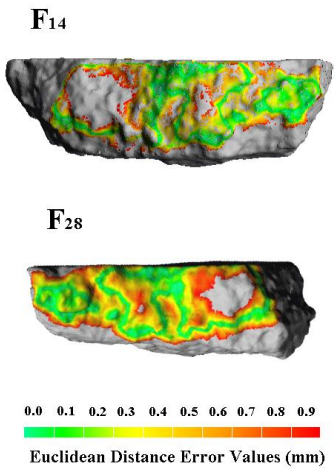

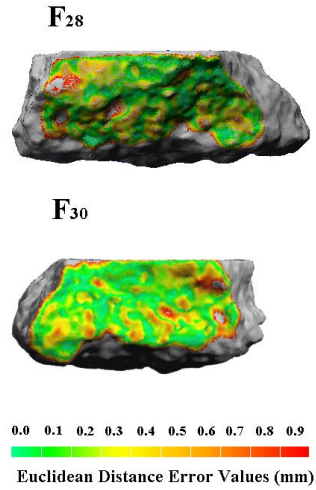

**Fig. 12:** Matching results and verification of island $I_{18}$. (a): The actual island formed by conservators according to the system suggestion. ($b_1$), ($c_1$): Pair-wise matches proposed by the system and the subsequent merge of the 3D representation of the two fragments at the matching position. ($b_2$), ($c_2$): Visualization of Euclidean distances between fragments surfaces at the matching positions depicted in ($b_1$) and ($c_1$) respectively.

# 6 Conclusion

In this paper, a novel methodology and a related system have been presented, for the reconstruction of fragmented objects having one near to plane surface. Five criteria have been developed to test if and exactly where two fragments match. The four of them describe a variety of geometric restrictions necessary for matching; they reject possible matching positions and they are applied sequentially. If all of them are satisfied at a considered relative position of the two fragments, then a fifth sufficient criterion is applied that takes into consideration the volume of a properly chosen gap between the two fragments. All these criteria and the way they are applied try to imitate the process the professional conservators follow. It also accounts for the unavoidable wear the fragments suffered and dynamically determines the extreme allowed geometric diversifications among actually matching fragments, by application of principles of Calculus of Variations.

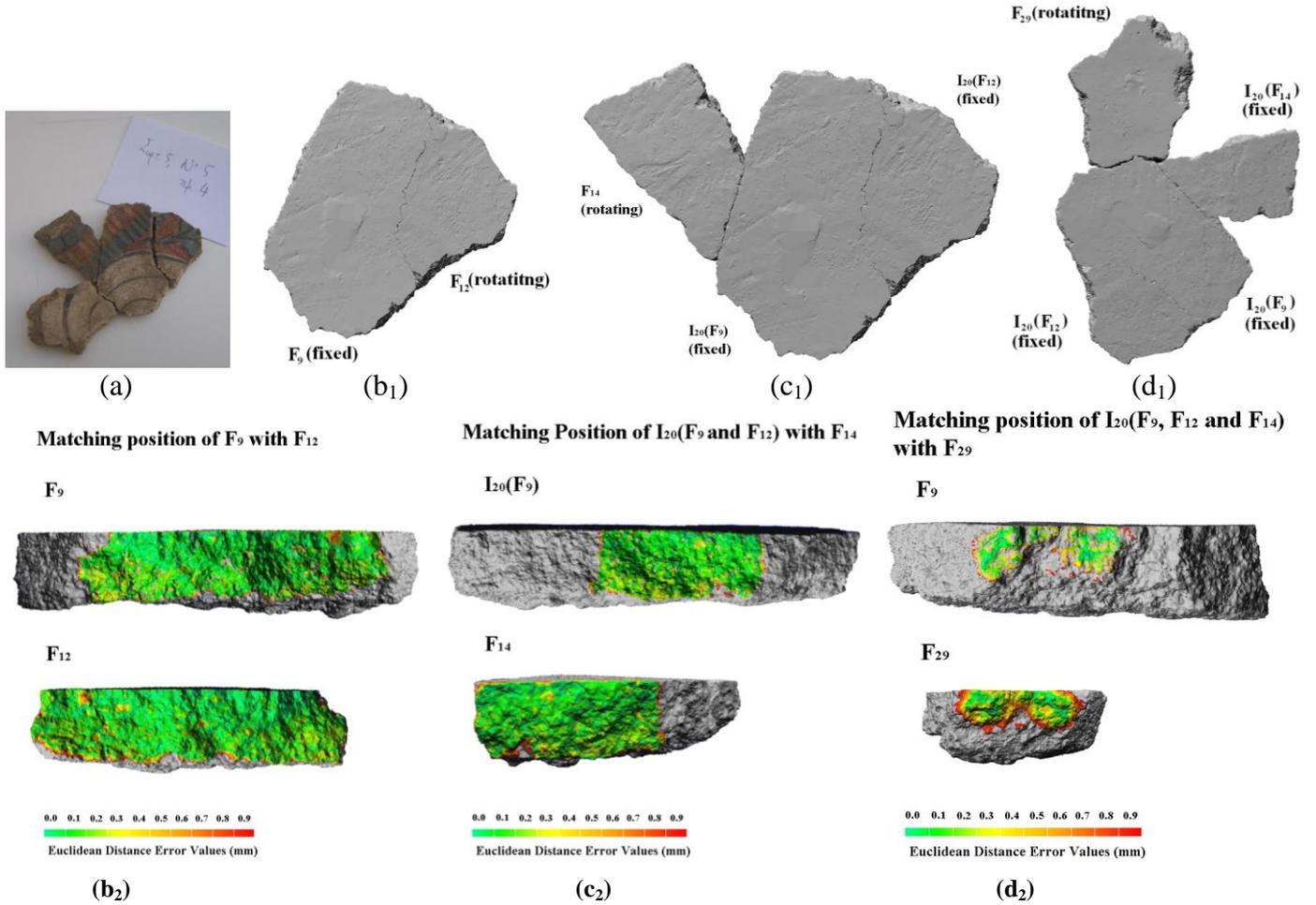

**Fig. 13:** Matching results and verification of island $I_{20}$ (a): The actual island formed by conservators according to the system suggestion. $(b_1)$, $(c_1)$, $(d_1)$: Pair-wise matches proposed by the system and the subsequent merge of the 3D representation of the two fragments at the matching position. $(b_2)$, $(c_2)$, $(d_2)$: Visualization of Euclidean distances between fragments surfaces at the matching positions depicted in $(b_1)$, $(c_1)$ and $(d_1)$ respectively.

The method has been applied to 2 cases: a) to the reconstruction of artificially broken flagstones with 100% success and b) to the reassembly of islands of fragments wall paintings belonging to the Mycenaic civilization ($14^{th}$-$13^{th}$ century B.C.) excavated at Tyrins, Greece. The scholars of the National Archaeological Museum of Greece, who perform reconstruction of fragmented archaeological objects manually, fully confirmed the results the introduced methodology offered. For example, see figures 7-15, which show both system results and the corresponding reconstruction, manually performed by the conservators. Moreover, the dedicated personnel of the museum did not find fragments islands additional to the ones offered by the system.

It would be really helpful in achieving the wall-paintings reconstruction to take into consideration the contour, thematic and color continuation between adjacent fragments [13], [12]. However, considering the form and wear of the illustrations appearing on the fragments, tackling this problem goes well outside the goals of the present work. The authors may deal with the problem in a future work.

The authors intend to apply the method and the system to a more large scale reconstruction, of prehistoric wall-paintings excavated at Tyrins and Mycenae. Finally, the authors are now extending the methodology, so that it can be applied to more general cases than the wall-paintings, such as the reconstruction of fragmented sculptures.

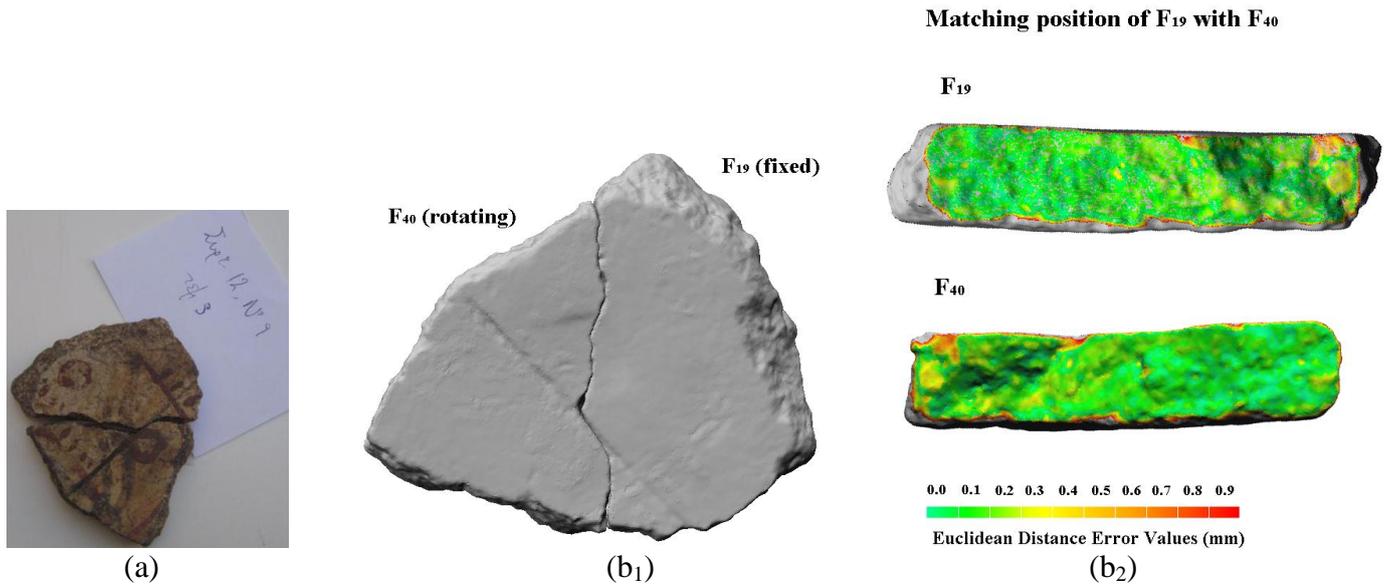

**Fig. 14:** Matching results and verification of island $I_{21}$. (a):The actual island formed by conservators according to the system suggestion. ($b_1$):Pair-wise matches proposed by the system and the subsequent merge of the 3D representation of the two fragments at the matching position. ($b_2$): Visualization of Euclidean distances between fragments surfaces at the matching positions depicted in ($b_1$).

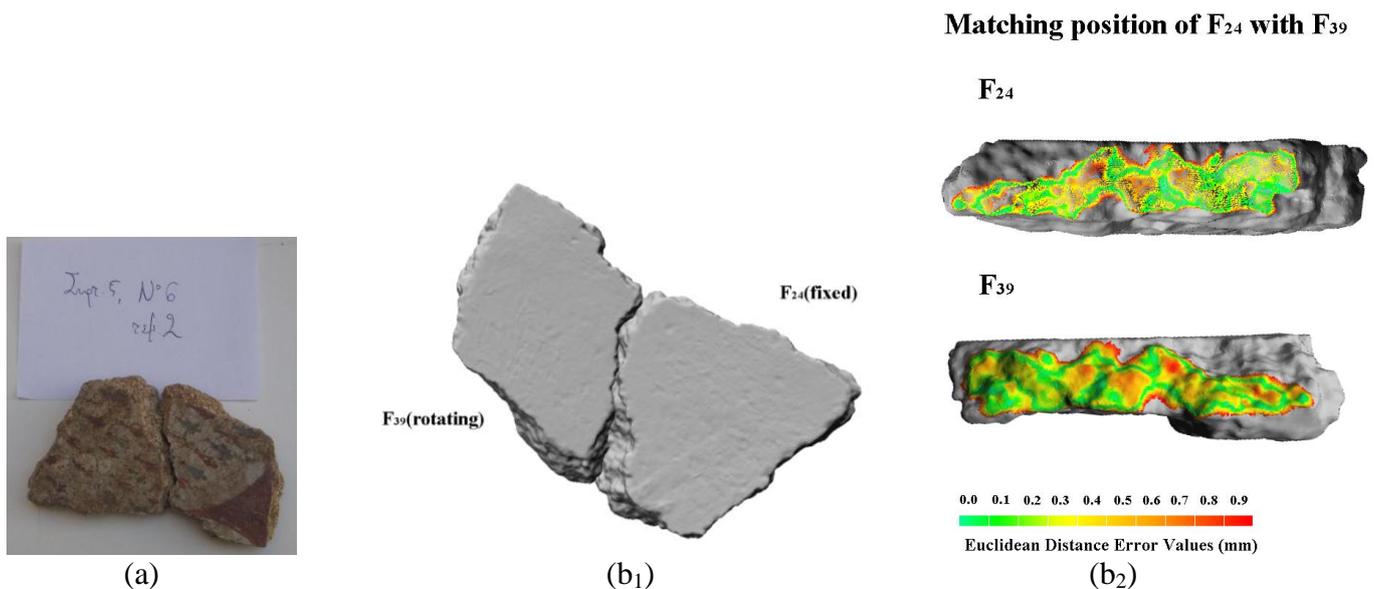

**Fig. 15:** Matching results and verification of island $I_{23}$. (a):The actual island formed by conservators according to the system suggestion. ($b_1$):Pair-wise matches proposed by the system and the subsequent merge of the 3D representation of the two fragments at the matching position. ($b_2$): Visualization of Euclidean distances between fragments surfaces at the matching positions depicted in ($b_1$).

# Appendix 1 – Determining near to plane surfaces on 3D fragments representation

We construct an auxiliary shape consisting of two parallel planes $\Delta_1$ and $\Delta_2$ having a distance $2\varepsilon$, where $\varepsilon$ is a properly chosen small quantity; let $\Delta$ be the middle, parallel to $\Delta_1$, $\Delta_2$, plane. If we imagine the situation where $\Delta$ coincides with the upper plane surface of the fragment, then for a proper choice of $\varepsilon$ all points of the fragment plane surface lie in the space between planes $\Delta_1$, $\Delta_2$; let $N^{US}$ be the number of these points. Moreover, if we consider the signed distance of all these $N^{US}$ points from $\Delta$, it is logical to expect that the mean value of these signed distances will be close to zero. On the contrary, it is quite logical to assume that in any other position of the auxiliary shape the number of points of the fragment surface lying between $\Delta_1$, $\Delta_2$ will be smaller than $N^{US}$ and/or that the mean value of their signed distances will be less close to zero. The value of the constant $\varepsilon$ that may guarantee this behavior, depends on the choppiness/undulation of the plain surface of the fragments, as well as on the resolution of the 3D digital representation and can be estimated by a trial and error method. A very satisfactory choice for all applications presented here seems to be $\varepsilon$ = resolution $2 \cdot 10^{-3}$. Hence, the plane surface of a fragment is determined via the following criterion:

We consider the space consisting of the three Euler angles $\delta\alpha, \delta\beta, \delta\gamma$ and the three translations $\delta x, \delta y, \delta z$ parallel to the axes $x,y,z$ respectively. For every point $(\delta\alpha, \delta\beta, \delta\gamma, \delta x, \delta y, \delta z)$ of this space, we perform the corresponding geometric transformations to the auxiliary shape fond by the three parallel planes $\Delta_1$, $\Delta_2$ and $\Delta$ and in the resulting position of the auxiliary shape we count the number $N$ of the fragment surface points that lie between $\Delta_1$, $\Delta_2$ as well as the mean value $\mu^D$ and the standard deviation $S^D$ of the signed distance of these points from $\Delta$. We determine the point $(\delta\alpha, \delta\beta, \delta\gamma, \delta x, \delta y, \delta z)$ for which $N$ is maximum and $\mu^D$ is less than $3.1 S^D / \sqrt{N}$. This point corresponds to a specific position of the auxiliary shape, for which the upper plane surface of the fragment optimally matches to $\Delta$.

The reason for the requirement $\mu^D < 3.1 S^D / \sqrt{N}$ lies on the fact that we plausibly suppose that the signed distances of the $N^{US}$ points from $\Delta$ follow a normal distribution with mean value 0 and that 99.9 % of the normal population satisfies this inequality. In practice this criterion is substantiated by application of a maximization algorithm in the space of points ($\delta\alpha,\delta\beta,\delta\gamma,\delta x,\delta y,\delta z$). Plain $\Delta_2$, namely the bottom plane of the optimally placed auxiliary shape (see fig 16), is defined to be the upper plane of the fragment.

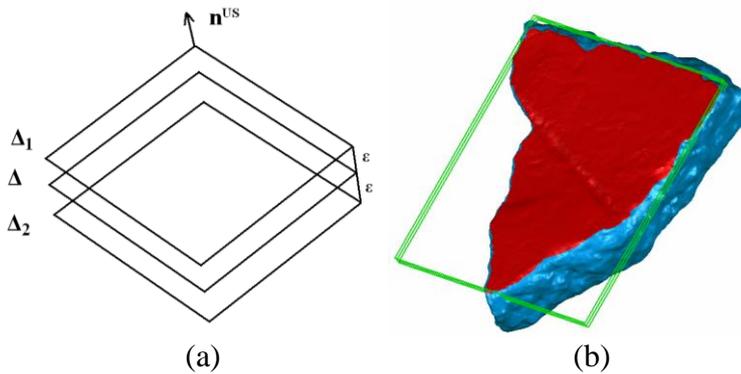

(a)        (b)

**Fig. 16:** The auxiliary shape and its use for the determination of fragments upper, near – to – plane, surface. (**a**): The configuration of the auxiliary shape as defined in Appendix 1. For each rotation of **n**$^{US}$ we count the points that lie between $\Delta_1$ and $\Delta_2$, and we calculate the mean signed distances of these points from $\Delta$.(**b**): The auxiliary shape in the orientation that offers maximal number of points lying between $\Delta 1$ and $\Delta 2$ (points in red) and signed distances from $\Delta$ close to 0.

## Appendix 2 – $\mathbf{A}^n$ is an Abelian Category

In order to determine a zero element for $\mathbf{A}^n$ we define a binary operator • in $D^{n-1} \times D^{n-1}$ so as for every $(E_1, E_2)$ and $(E_1', E_2')$ in $D^{n-1} \times D^{n-1}$ we have $(E_1, E_2) \bullet (E_1', E_2') = \left(E_1 \cap (E_2 \sqcup E_1'), E_1' \cap (E_1 \sqcup E_2')\right)$. This binary operator maps any collection of elements in the form $(E_i, \emptyset) \in D^{n-1} \times D^{n-1}$ to elements in the form $\left(E_i \cap E_j, E_i \cap E_j\right)$. Then, each pair $\left(E_i, E_j\right)$ is uniquely determined by $(E_i, E_i) \bullet \left(E_j, E_j\right)$ and is uniquely mapped into elements of the form $(E_i, \emptyset)$ by $\left(E_i, E_j\right) \bullet (\emptyset, E_i)$. Thus operator • allows for recycling between $D^{n-1} \times \emptyset \xrightarrow{\bullet} D^{n-1} \times D^{n-1} \xrightarrow{\bullet} D^{n-1} \times \emptyset$ by a unique sequence thus making $\emptyset$ initial and terminal object for $D^{n-1} \times D^{n-1}$ at the same time, i.e. a zero object. For the objects in $D^n$, by applying • to the binary products in $D^n$, proceeding as in the case of $D^{n-1} \times D^{n-1}$, we can construct a unique sequence of morphisms for each $D \in D^n$ $\emptyset \to (D, \emptyset) \xrightarrow{(\emptyset,D)\bullet} (\emptyset, \emptyset) \to \emptyset$, thus obtaining $\emptyset$ as the zero object for $D^n$. Concerning $\mu^n$, since it is the class of measures of $D^n$, it has elements in the quotient $(D^n \to \mathbb{R})/\text{im }m$. This means that for any measure $\mu \in \text{im }m$ and any 2 elements $f_1, f_2$ of $\mu^n$ in the equivalence class $[\mu]$ there exist both morphisms $f_1 \to f_2$ and $f_2 \to f_1$. So, since $\mu = 0$ is in one-to-one correspondence with $\emptyset$, by defining two epimorphisms $\mu \to f$ and $\mu \to f \cdot \mu$, for an $f \in \mu^n$, there exist both morphisms $0 \to f$ and $f \to 0$. Since $\mu \to f$ is an epimorphism these 2 morphisms are unique. Thus, $\emptyset$ is also the zero element of $\mu^n$. We note that the fact that endomorphisms in $\mu^n$ are epimorphisms is a consequence of the fact that $m$ is an epimorphism which implies that $\text{Hom}(\text{im }m, \mu^n) \to \text{Hom}(D^n, \mu^n)$ is an injection making $\text{im }m \to \mu^n$ an epimorphism. Thus $\mathbf{A}^n$ has $\emptyset$ as its zero object.

Moreover, we can define binary products for all elements of $\mathbf{A}^n$ in the form of Cartesian products by simply distributing the morphisms to the pairs that constitute the Cartesian products and include canonical projections to the morphisms of $\mathbf{A}^n$. Concerning the co-products of $\mathbf{A}^n$, for the elements of $D^n$ a co-product can be defined by the disjoint union of sets since for an inclusion $D \xrightarrow{i} D \sqcup D'$ of a $D \in D^n$ and for some $D' \subseteq D$, $m(D \sqcup D') = m(D) + m(D') - m(D \cap D') = m(D)$. The co-products of the elements of $D^{n-1} \times D^{n-1}$, can be constructed using operator • and determining its effect on $t$. Namely, we exploit the identity that for 2 boundary elements $(E_1, E_2) \in D^{n-1} \times D^{n-1}$ their subset $E_1 \cap E_2$ participates only to the boundary of a domain formed through $t$. Thus, for some element $E \in D^{n-1}$ so that $E_1 \sqcup E$ and $E_2 \sqcup E$ are still bounded domains, it holds that $t(E_1 \sqcup E, E_2 \sqcup E) = t(E_1, E_2) \sqcup E$, thus offering that $t((E_1, E_2) \bullet (E'_1, E'_2)) = t(E_1 \cap E_2, E'_1 \cap E'_2) \sqcup (E_1 \cap E'_1)$. Then a co-product of $D^{n-1} \times D^{n-1}$ can be defined to be operator •, as each pair $(E_1, E_2)$ uniquely corresponds through • to $(E_1, E_2) = (E_1, E_1) \bullet (E_2, E_2)$ and $t(E_1, E_2) = t(E_1, E_2) \sqcup E_1 \cap E_2$, because $E_1 \cap E_2 \subseteq E_1, E_2$.

## Acknowledgements

The warmest thanks to C. Drettas for offering the 3D scanner and for his valuable support to the scanning process. We, also, warmly thank the archaeologist K. Paschalidis and the conservator E. Velalopoulou for their valuable assistance in the storerooms of the Prehistoric Collection of the National Archaeological Museum.

## References

[1] Besl, P.J. and McKay, H.D. (1992). A method for registration of 3-D shapes. IEEE Trans. on Pattern Analysis and Machine Intelligence, 14(2), 239-256.

[2] Biswas, A., Bhowmick, P., and Bhattacharya, B.B. (2005). Reconstruction of torn documents using contour maps. IEEE Int. Conf. on Image Processing, 3, 517-520.

[3] Brown, B.J., Toler-Fraklin, C., Nehab, D., Burns, M., Dobkin, D., Vlachopoulos, A., Doumas, C. Rusinkiewicz, C. and Weyrich, T. (2008). A system for high-volume acquisition and matching of fresco fragments: reassembling Theran wall paintings. ACM Transactions on Graphics, 27( 3).

[4] da Gama Leitao, H.C., Stolfi, J. (2002). A multiscale method for the reassembly of two-dimensional fragmented objects. IEEE Trans. on PAMI, 24(9), 1239-1251.

[5] Huang, Q.X., Flöry, S., Gelfand, N., Hofer, M. and Pottmann, H. (2006). Reassembling fractured objects by geometric matching. ACM Transactions on Graphics, 25(3), 569-578.

[6] Igwe P.C. and Knopf G.K. (2006). 3D Object Reconstruction Using Geometric Matching. IEEE Proceedings of the Geometric Modeling and Imaging – New Trends, 16-18, 9-14.

[7] Linnainmaa, S., Harwood, D., and Davis, L. S. (1988). Pose determination of a three-dimensional object using triangle pairs. IEEE Transactions on Pattern Analysis and Machine Intelligence, 10(5), 634–647.

[8] Papaioannou, G., Karabassi, E.A., and Theoharis, T. (2002). Reconstruction of Three-Dimensional Objects through Matching of Their Parts. IEEE Trans. on Pattern Analysis and Machine Intelligence, 24(1), 114-124.

[9] Papaioannou, G. and Karabassi, E.A. (2003). On the automatic assemblage of arbitrary broken solid artefacts. Elsevier Image and Vision Computing, 21, 401-412.

[10] Papaodysseus, C. Panagopoulos, T. Exarhos, M. Triantafillou, C. Fragoulis, D. Doumas, C. (2002). Contour-shape based reconstruction of fragmented, 1600 BC wall-paintings. IEEE Transactions on Signal Processing, 50(6), 1277-1288.

[11] Papaodysseus, C., Arabadjis, D., Panagopoulos, M., Rousopoulos, P., Exarhos , M. and Papazoglou, E. (2008). Automated reconstruction of fragmented objects using their 3D representation - application to important archaeological finds. IEEE Proc. of ICSP 08, 769-772.

[12] Papaodysseus, C. Panagopoulos, T. Exarhos, M. Triantafillou, C. Fragoulis, D. Doumas, C. (2008). Image and Pattern Analysis of 1650 B.C. Wall Paintings and Reconstruction. IEEE Transactions on Systems, Man and Cybernetics, 38(4), 958-965.

[13] W. Puech, A.G. Bors, I. Pitas, J.-M. Chasssery, "Projection Distortion Analysis for Flattened Image Mosaicing from Straight Uniform Generalized Cylinders," Pattern Recognition, vol. 34, no. 8, pp. 1657-1670, Aug. 2001

[14] Shogo, Y., Shohei, K., Satoshi, K. and Hidenori, I. (2005). An Earthenware Reconstruction Method Based on the Matching of Both Contour Curve and Color Pattern on the Surface of Potsherds. Journal of the Institute of Image Electronics Engineers of Japan, 34(2), 126-133.

[15] de Smet, P. (2008). Reconstruction of ripped-up documents using fragment stack analysis procedures, Elsevier Forensic Science International, 176(2-3), 124-136.

[16] Stanco F., Tenze L., Ramponi G., and de Polo A., (2004). Virtual restoration of fragmented glass plate photographs. 12th IEEE Medit. Electrotechnical Conf., 1, 243-246.

[17] Ukovic, A., and Ramponi, G. (2008). Feature extraction and clustering for the computer-aided reconstruction of strip-cut shredded documents. J. Electron. Imaging, 17(1).

[18] Ukovic, A., Ramponi, G., Doulaverakis, H., Kompatsiaris, Y., and Strintzis, M.G. (2004). Shredded document reconstruction using MPEG-7 standard descriptors. 4th IEEE Int. Symp. on Signal Processing, 334-337.

[19] Ucoluk, G., Toroslu, I.H. (1999). Automatic reconstruction of 3D surface objects. Elsevier Computers & Graphics, 23(4), 573-582.

[20] Valkenburg, R. J. and.McIvor, A. M. (1998). Accurate 3D measurement using a structured light system, Elsevier Image Vision and Computing, 16(2), 99-110.

[21] Willis, A., Orriols, X., Cooper, D.B. (2003). Accurately Estimating Shred 3D Surface Geometry with Application to Pot Reconstruction. Proceedings on Computer Vision and Pattern Recognition, 1-7.

[22] Winkelbach, S., Rilk, M., Schönfelder, C. and Wahl, F. M. (2004). Fast random sample matching of 3D fragments. Lecture Notes in Computer Science: 3175. Pattern recognition, 26th DAGM symposium, Berlin: Springer, 129–136.

[23] Winkelbach, S. and Wahl, F.M. (2008). Pairwise Matching of 3D Fragments Using Cluster Trees. Springer Int. J. Computer Vision, 78(1), 1-13.

[24] Zhu, L., Zhou, Z. and Hu, D. (2008). Globally Consistent Reconstruction of Ripped-Up Documents. IEEE Trans. on PAMI, 30(1), 1-13.